\documentclass{article}

\usepackage{microtype}
\usepackage{graphicx}
\usepackage{subcaption}
\usepackage{booktabs} 
\usepackage{dblfloatfix} 

\usepackage{hyperref}

\usepackage[accepted]{icml2026}

\usepackage{amsmath}
\usepackage{amssymb}
\usepackage{mathtools}
\usepackage{amsthm}

\usepackage[capitalize,noabbrev]{cleveref}

\theoremstyle{plain}

\theoremstyle{definition}

\theoremstyle{remark}

\usepackage[textsize=tiny]{todonotes}

\icmltitlerunning{Design-CP}

\begin{document}

\twocolumn[
  \icmltitle{Design-CP: Context Parallelism for Design of Protein Nanoparticles}




  \icmlsetsymbol{equal}{*}
  \begin{icmlauthorlist}
    \icmlauthor{Lorenzo Tarricone}{yyy,comp}
    \icmlauthor{Helen E. Eisenach}{zzz,xxx}
    \icmlauthor{Aiko Muraishi}{zzz,xxx,www}
    \icmlauthor{Charlotte M. Deane}{yyy}
  \end{icmlauthorlist}

  \icmlaffiliation{yyy}{Department of Statistics, University of Oxford, Oxford, UK}
  \icmlaffiliation{comp}{Ellison Institute of Technology Oxford, Oxford, UK}
  \icmlaffiliation{xxx}{Department of Biochemistry, University of Washington, Seattle, USA}
  \icmlaffiliation{zzz}{Institute for Protein Design, University of Washington, Seattle, USA}
  \icmlaffiliation{www}{Paul G. Allen School of Computer Science and Engineering, University of Washington, Seattle, USA}

  \icmlcorrespondingauthor{Charlotte Deane}{deane@stats.ox.ac.uk}

  \icmlkeywords{Machine Learning, BioML, Protein Design, Parallelism, Protein Nanoparticles}

  \vskip 0.3in
]



\printAffiliationsAndNotice{}  

\begin{abstract}
Many all-atom generative protein models can in principle design large multimeric complexes by jointly modelling all chains, but their quadratic token- and atom-pair representations quickly exceed single-GPU memory as the number of chains and residues modelled grows. We introduce \emph{Design-CP}, two context-parallel (CP) inference strategies for RFdiffusion~3 (1D row-sharding and 2D grid sharding with ring attention) that distribute the quadratic activations across a multi-GPU mesh while preserving pretrained weights. We characterise their scaling when sampling icosahedral assemblies, showing that the maximum feasible asymmetric subunit (ASU) size grows with the expected square-root trend in GPU count and that 2D sharding achieves better wall-clock scaling. Moreover, we show how strong point-group symmetry constraints make CP usable out of the box for end-to-end, all-atom design of icosahedral nanoparticles, yielding favourable \emph{in silico} structural and interface metrics. Finally, we demonstrate octahedral nanoparticle design on a small cluster of workstation-grade 16\,GB GPUs, illustrating how Design-CP can be a practical path towards democratising large-assembly protein design.
\end{abstract}

\section{Introduction}
\label{sec:introduction}

Deep learning is transforming computational protein design from a predominantly physics-based endeavour into a data-driven discipline. Families of structure prediction models such as AlphaFold~\cite{jumperHighlyAccurateProtein2021, abramsonAccurateStructurePrediction2024}, RoseTTAFold~\cite{baekAccuratePredictionProtein2021, baek_efficient_2023, corley_accelerating_2025} and Boltz~\cite{wohlwend_boltz-1_2024, passaroBoltz2AccurateEfficient2025} now achieve near-experimental accuracy on many single-chain targets. In parallel, a rapidly expanding family of denoising-based generative design frameworks including RFDiffusion \citep{watsonNovoDesignProtein2023, butcher_novo_2025}, Chroma \citep{ingrahamIlluminatingProteinSpace2023}, Genie~ \citep{lin_generating_2023, lin_out_2024}, and Proteina \citep{geffner_proteina_2025, geffner_-proteina_2025, didi_scaling_2026} 
enable the \emph{de novo} creation of proteins with prescribed structural and functional properties. These advances have already yielded a tangible impact across diverse application domains. In therapeutic design alone, examples include de novo minibinders against therapeutically relevant targets such as
bioactive peptide hormones~\cite{vazquez_torres_novo_2024} and bacterial toxins~\cite{ragotte_novo_2025} as well as the de novo design of epitope-targeted antibodies, from
diffusion-based co-design of CDR sequence and structure on a fixed
framework~\cite{luo_antigen-specific_2022} to atomically accurate
in-silico design of VHHs and scFvs~\cite{bennett_atomically_2026}. \\

The success of these methods motivates scaling such generative tools to larger and more biologically complex targets, but this requires the ability to reliably design \emph{multimeric} protein complexes. In nature, the majority of proteins carry out their functions not as isolated monomers but as oligomeric assemblies, such as homodimers, heteromeric complexes, and higher-order symmetric architectures \cite{goodsell_structural_2000, marsh_structure_2015}. Designing symmetric assemblies \emph{de novo} could unlock applications ranging from biomolecular machines inspired by rotary motors such as ATP synthase \cite{courbet_computational_2022} to vaccine scaffolds inspired by viral capsids \cite{butterfield_evolution_2017, marcandalliInductionPotentNeutralizing2019a, walls_elicitation_2020}. The computational design of such assemblies has so far relied on rigid-body docking of independently-designed oligomers~\citep{king_computational_2012, bale_accurate_2016, hsia_design_2016, sheffler_fast_2023}, a paradigm that recent ML-era pipelines~\citep{ de_haas_rapid_2024, haas_sequence_2025, haas_novo_2026} have refined but not fundamentally replaced. Crucially, this reliance on docking is often a practical workaround rather than a modelling choice: end-to-end all-atom generators exist, but they struggle to fit whole assemblies in memory when many subunits must be modelled jointly. \\

Recent all-atom generative models such as RFDiffusion 3~\cite{butcher_novo_2025} (RFD3), which is inspired by the AlphaFold 3 architecture~\cite{abramsonAccurateStructurePrediction2024} (AF3), can in principle generate multimeric structures by jointly modelling all chains with an atomistic level of precision and designing proteins with predefined point-group symmetries. However, the underlying architecture maintains pairwise representations whose memory cost scales quadratically with the number of tokens~$I$ (and atoms~$L$). For large protein assemblies, $I$ and $L$ grow linearly with the number of chains modelled, causing the $\mathcal{O}$($I^{2}$) and $\mathcal{O}$($L^{2}$) pairwise feature tensors and related quadratic intermediates to exceed the memory capacity of a single GPU. This practical bottleneck heavily limits the size of what can be designed, particularly when modelling symmetric protein assemblies. This single-device ceiling, however, contrasts with broader trends in computational infrastructure: while single-device memory capacity has grown only incrementally, access to multi-GPU clusters is now routine for both academic and industrial research groups. Accordingly, partitioning large transformer workloads across such clusters has become standard practice in adjacent fields such as large language modelling, both for training~\cite{li_sequence_2022} and inference~\cite{pope_efficiently_2022}.\\

We implement and compare two context-parallel (CP) inference strategies for RFD3, which we call \emph{Design-CP}. The first, a \emph{1D} scheme, stripes the pair representation across $P$ GPUs along a single axis; the second, a \emph{2D} scheme following Fold-CP~\cite{lin_fold-cp_2026}, tiles it over a $\sqrt{P}\times\sqrt{P}$ device grid with ring attention. Both shard the dominant quadratic memory cost while preserving numerical equivalence with single-GPU inference. We evaluate the two schemes on large point-group-symmetric assemblies, including icosahedral and octahedral nanoparticles, and show that symmetry constraints sharpen practical sample quality and make end-to-end all-atom sampling tractable on modest multi-GPU setups without additional training or fine-tuning.

\paragraph{Contributions.} 

Our main contributions are:
\begin{itemize}
    \item \textbf{Design-CP: context-parallel inference for RFD3 with strong scaling.} We introduce two CP schemes for RFD3 inference (a lightweight \emph{1D row-sharding} strategy and a \emph{2D grid} strategy based on Fold-CP~\cite{lin_fold-cp_2026}), and we characterise their memory ceilings and wall-clock scaling when sampling large symmetric icosahedral assemblies.
    \item \textbf{Symmetry makes CP usable out of the box for icosahedral design.} We show that imposing strong point-group symmetry constraints sharpens practical sample quality beyond the native training crop and makes end-to-end, all-atom generation of icosahedral nanoparticles tractable without retraining or fine-tuning.
    \item \textbf{Octahedral design on small GPUs.} We demonstrate that the same approach enables \emph{de novo} design of octahedral nanoparticles on a small cluster of workstation-grade GPUs, showcasing a workable route to making large-assembly protein design broadly accessible.
\end{itemize}

\section{Related Work}
\label{sec:related_work}

Two lines of prior work are directly relevant to Design-CP: (i) methods that reduce the memory cost of AF3-class architectures, which we build on technically, and (ii) computational pipelines for designing large symmetric protein assemblies, where Design-CP aims to make a methodological contribution. For the first, we briefly cover IO-efficient attention on a single device, while devoting most of this section to distributed parallelism for structure models. For the second, we situate Design-CP within the broader landscape of protein nanoparticle design methods.

\paragraph{IO-efficient attention.} FlashAttention \cite{dao_flashattention_2022, dao_flashattention-2_2023}
computes exact attention with memory linear in sequence length ($\mathcal{O}(I)$ or $\mathcal{O}(L)$ in our notation) by tiling queries, keys, and values into SRAM-resident blocks while maintaining running
softmax statistics \citep{milakov_online_2018}, building on the earlier observation that attention admits a linear-memory implementation \citep{rabe_self-attention_2022}. For architectures without pair representations, this suffices and motivates its use in protein language models like ESM3 \citep{hayesSimulating500Million2025}, which represent sequence, structure, and function as discrete tokens and condense pairwise geometry into a single SE(3)-invariant geometric-attention block at the input. Flash‑IPA \citep{liu_flash_2025} and FlashBias \citep{wu_flashbias_2025} extend this idea to pair‑biased attention, such as the one present in AlphaFold‑3 Pairformer, by re-expressing geometric and pairwise bias terms via additional low‑rank features concatenated into the query and key projections. For complex learned pair biases, these methods generally require training auxiliary parameters and are not a weight‑preserving drop‑in for an arbitrary pretrained model. These approaches are orthogonal to Design-CP: they reduce the per-block memory footprint on a single device, while we shard the persistent quadratic activations across devices, and the two could in principle be composed.

\paragraph{Distributed parallelism for structure models.} Among the standard parallelism axes (data, tensor, pipeline, expert, activation), \emph{context parallelism} uniquely shards activations along the sequence dimension of every layer, which makes it a natural fit for the $\mathcal{O}(I^2)$ pair tensor of AlphaFold-class models. FastFold \citep{cheng_fastfold_2023} introduced Dynamic Axial Parallelism (DAP) for AlphaFold~2, replicating parameters on every device and sharding activations along a single sequence axis at a time; ScaleFold \citep{zhu_scalefold_2024} adopted DAP and scaled to 2048 H100s for training. As the Evoformer interleaves row- and column-wise attention over the MSA track, DAP must insert an all-to-all communication step whenever the active axis flips, incurring six all-to-all redistributions per Evoformer block at inference, together with one \textsc{AllGather} in the outer-product-mean and two in the triangular updates~\citep{cheng_fastfold_2023}. These collectives contribute non-trivially to inference latency at scale and increase transient memory relative to the steady-state ($\mathcal{O}(I^2/P)$) sharded pair representation.

Fold-CP \citep{lin_fold-cp_2026} generalises axis sharding to a full two-dimensional context-parallel strategy for AF3-class models (implementing it for Boltz-2~\citep{passaroBoltz2AccurateEfficient2025}), extending Ring Attention~\citep{liu_ring_2023} into a Cannon-style 2D ring tailored to dense triangular updates, while window-batched atom attention is handled by a complementary shardwise kernel that keeps each window's attention local to its rank. The pair tensor is tiled over a $\sqrt{P}\times\sqrt{P}$ device grid, so that rank $(r,c)$ holds the block indexed by residue ranges $[r\cdot I/\sqrt{P}, (r{+}1)\cdot I/\sqrt{P}]\times [c\cdot I/\sqrt{P}, (c{+}1)\cdot I/\sqrt{P}]$. Triangle attention, triangle multiplication, attention-with-pair-bias, pair-weighted-averaging, and outer-product-mean are each reformulated as ring algorithms in which $\mathbf{K}$ and $\mathbf{V}$ shards (and, where required, triangular biases and masks) circulate between neighbouring devices while a numerically-stable tiled softmax merges partial outputs without ever materialising the full $I\times I$ attention on any rank. The resulting steady-state pair memory per device is ($\mathcal{O}(I^2/P)$), matching 1D axis-sharded DAP. However, under 1D sharding, triangular updates typically require collectives over all ($P$) ranks (e.g., to obtain the necessary key/value or bias shards along the active axis), which can inflate the transient working-set memory during attention/multiplication beyond the steady shard. In Fold‑CP’s 2D tiling, collectives are restricted to a single row or column subgroup of size ($\sqrt{P}$). This reduces communication volume and confines the transient working-set memory growth in triangular updates (e.g., gathered/circulated K/V shards, triangle biases, masks, and other pair-like intermediates) to ($\sqrt{P}$)-sized groups, rather than requiring global ($P$)-way collectives.

To the best of our knowledge, context parallelism has so far been developed and evaluated exclusively for structure \emph{prediction}. We take this as motivation to apply the same techniques to generative \emph{design}, and adopt RFdiffusion~3 \citep{butcher_novo_2025} as our target. Our 2D scheme is a direct port of Fold-CP \citep{lin_fold-cp_2026}. At the same time, we observe that RFD3 has no MSA processing or triangular operations, which makes a much lighter 1D row-sharded scheme practical as well. We implement both and compare them on symmetric-design tasks.

\paragraph{Computational design of protein nanoparticles.}
\citet{king_computational_2012} introduced the modern \emph{dock-and-design} recipe in Rosetta: pre-existing oligomeric building blocks with compatible point-group symmetry are docked as rigid bodies along the rotational axes of the target architecture, sampling only the radial displacement $r$ and axial rotation $\omega$, and a new low-energy interface is then sequence-designed between them. Extending the pipeline to nanoparticles of multiple components~\citep{king_accurate_2014} enabled scaling to megadalton-scale icosahedral assemblies such as the 120-subunit I53-50~\citep{bale_accurate_2016} and the hyperstable 60-subunit I3-01~\citep{hsia_design_2016}. Recent work has progressively replaced individual modules of this pipeline in favour of ML-based methods: ProteinMPNN substitutes for Rosetta in interface design~\citep{de_haas_rapid_2024}; AlphaFold2 predictions of thermophilic homologs supply building blocks in place of experimental structures~\citep{haas_sequence_2025}; and RFdiffusion-generated \emph{de novo} oligomers now serve as the building-block library~\citep{haas_novo_2026}. Crucially, all of these methods still rely on a rigid-body docking step over pre-computed, independently generated oligomers. A natural next step is to try to model the entire assembly with a single generative network, but the per-device memory footprint of representing a megadalton-scale complex at full-atom resolution currently makes joint generative design infeasible on a single GPU.

  Design-CP removes this single-GPU memory barrier, enabling RFdiffusion~3 to jointly denoise all atoms of large multimeric proteins in a single trajectory: the first end-to-end, all-atom generative design of symmetric protein assemblies that models the full set of inter-ASU interactions without an intermediate docking step.

\begin{figure*}[!t]
  \centering
  \includegraphics[width=\textwidth]{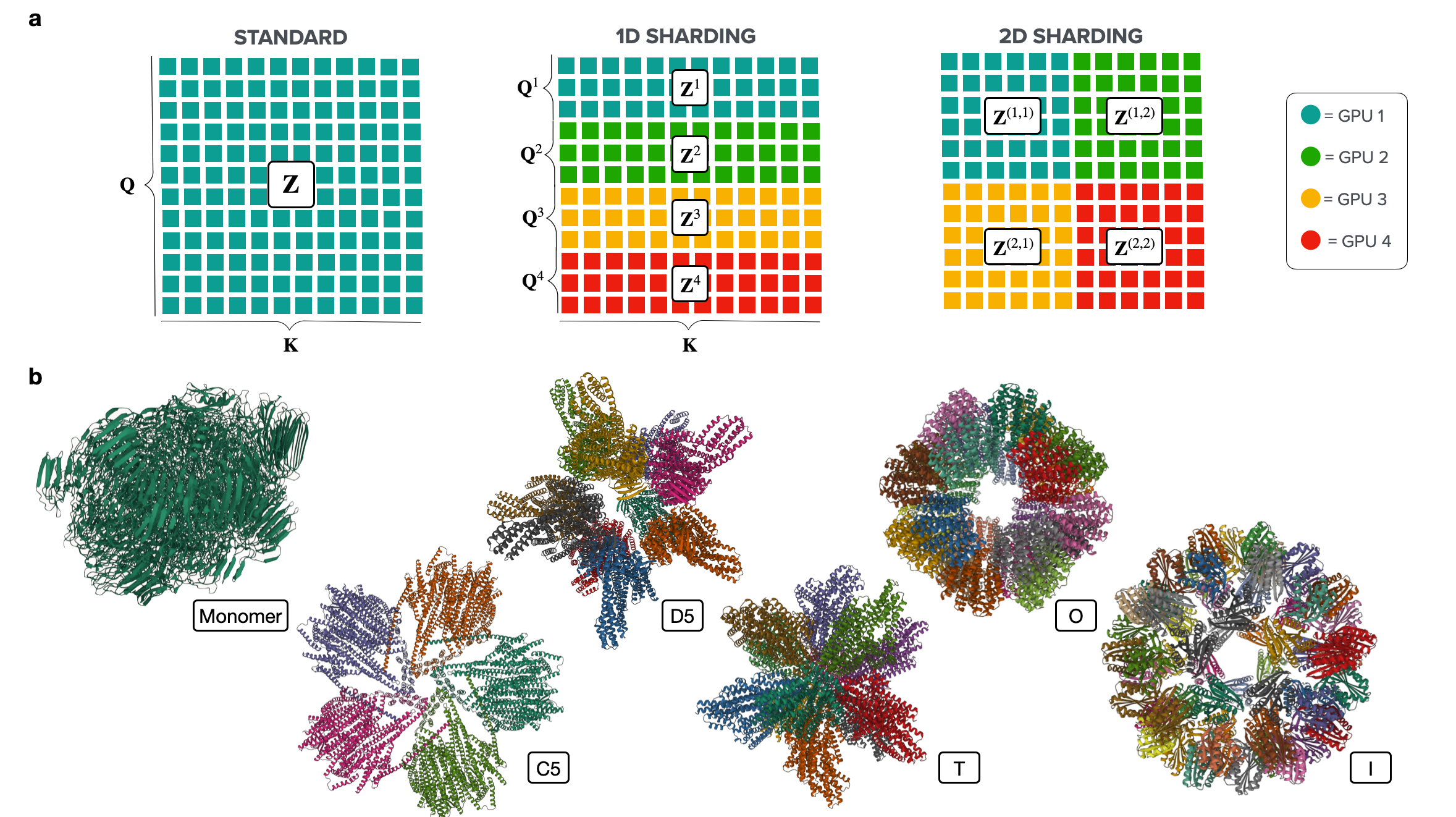}
  \caption{\textbf{Symmetric design with Design-CP}.\textbf{a}, Schematic depiction of the sharding techniques implemented in Design-CP. Every square represents a sub-tensor of the self-attention matrix, and its colour represents its assigned device. 1D sharding partitions the queries across different GPUs, where they are used to calculate cross-attention against all keys. 2D sharding partitions both queries and keys and uses ring attention to compute attention scores on the fly. \textbf{b},\textbf{ Qualitative comparison of large-design samples.} Designs of a 10800 amino acid protein using different symmetries. Designs that exceed the native crop limits can show visible degradation when sampled without additional structure constraints; imposing a strong symmetry prior mitigates this effect by reducing the effective design space.}
  \label{fig:large-monomer-quality}
\end{figure*}

\section{Methods}
\label{sec:methods}

\subsection{RFDiffusion 3 preliminaries and notation}
\label{sec:prelims}

RFDiffusion 3 (RFD3) jointly models $I$ tokens (each token representing, for example, a single residue or the heavy atom of a small molecule) together with $L$ atoms. Tokens carry a single-track tensor $\mathbf{S}\in\mathbb{R}^{I\times c_s}$ and a pair representation $\mathbf{Z}\in\mathbb{R}^{I\times I\times c_z}$; atoms carry a single-track $\mathbf{A}\in\mathbb{R}^{L\times c_\text{atom}}$ and an analogous pair representation $\mathbf{P}\in\mathbb{R}^{L\times L\times c_\text{atompair}}$. Within every token-level attention block, a learned projection of $\mathbf{Z}$ produces a per-head pair bias $\mathbf{B}\in\mathbb{R}^{I\times I\times H}$ that is added to the attention logits before the softmax, $\operatorname{softmax}(\textbf{QK}^{\top}/\sqrt{d}+\mathbf{B})$. Atom-level attention is \emph{sparse} in computation: each query atom attends to a budget of $k$ neighbours assembled from a small set of atoms close in sequence together with the spatially closest atoms, with usually $k\!\ll\!L$. The relevant slice of $\mathbf{P}$ is gathered at those $k$ indices and then projected to the per-head bias, so the attention computation itself is $\mathcal{O}(L\,k)$. The storage cost of $\mathbf{P}$, however, remains quadratic, and the dense $[L,L,c_\text{atompair}]$ tensor is the dominant atom-level memory consumer at the scales we target.\footnote{The pre-existing RFD3 inference codebase also exposes an optional low-memory mode based on a chunked pairwise embedder that constructs $\mathbf{P}$ on the fly at the $k$ kNN indices, avoiding the dense materialisation; this is orthogonal to Design-CP and we describe it in Appendix~\ref{app:rfd3-bg}.} In this work, we successfully partitioned the $\mathcal{O}(I^{2})$ pair track $\mathbf{Z}$ and the $\mathcal{O}(L^{2})$ pair track $\mathbf{P}$ across $N$ GPUs while never communicating full pair tensors. A fuller description of the five-stage RFD3 architecture and of the $\mathbf{P}$--kNN interaction is deferred to Appendix~\ref{app:rfd3-bg}.

When designing with a pre-defined point symmetry, the diffusion model is always run on the entire complex at every denoising step. For a configurable fraction of the trajectory (default 90\%), RFD3 then \emph{resymmetrises} its prediction by extracting the coordinates of a single asymmetric subunit (ASU) and generating the remaining copies by applying the point-group operations. The resulting resymmetrised complex is the state that is fed into the next iteration of the sampling loop. Additional details on the symmetrisation procedure are deferred to Appendix~\ref{app:symmetry}.

\subsection{1D row-sharding}
\label{sec:1d}

Our first scheme, implemented in PyTorch with NCCL and without custom kernels or DTensor machinery, partitions the query dimension of every $I\times I$ and $L\times L$ operation across $P$ GPUs (\cref{fig:large-monomer-quality}a). GPU~$p$ materialises only its row stripe of the pair track:
\begin{equation}
\mathbf{Z}^{(p)} \in \mathbb{R}^{I_p \times I \times c_z}, \qquad
I_p = \lfloor I/P \rfloor + \mathbb{I}[p < I \bmod P],
\label{eq:stripe}
\end{equation}
where $\mathbb{I}[\,\cdot\,]\in\{0,1\}$ denotes the indicator function of its predicate. When $I$ is not divisible by $P$, the floor $\lfloor I/P\rfloor$ leaves a remainder of $I\bmod P$ elements that must still be assigned. We absorb this remainder by giving the first $I\bmod P$ ranks one additional row each: rank $p$ receives the extra row precisely when $p < I\bmod P$, which is what the indicator encodes. By construction $\sum_{p}I_p = I$, no element is dropped, and chunk sizes differ by at most one across GPUs, so the load imbalance per attention block is bounded by a single row regardless of $P$. The pair tracks $\mathbf{Z}^{(p)}$ and the self-conditioning distogram $\mathbf{D}^{(p)}_\text{self}$ are in this way \emph{never} gathered to their full $[I,I]$ shape.

Every self-attention block is reformulated as a cross-attention: GPU~$p$ projects queries from its stripe $\mathbf{Q}^{(p)}=f_Q(\mathbf{S}^{(p)})\in\mathbb{R}^{I_p\times H\times d}$, while keys and values are projected from the \emph{full} single-track $\mathbf{S}$, which is replicated on every GPU. The pair bias is drawn from the local stripe, $\mathbf{B}^{(p)}=f_B(\mathbf{Z}^{(p)})\in\mathbb{R}^{I_p\times I\times H}$, and
\begin{equation}
\operatorname{Attn}^{(p)} = \operatorname{softmax}\!\left(
  \mathbf{Q}^{(p)} \mathbf{K}^\top/\sqrt{d} + \mathbf{B}^{(p)}
\right) \mathbf{V}
\label{eq:cross_attn}
\end{equation}
A single \textsc{AllGather} over the 1D track reconstructs $\mathbf{S}=\bigoplus_{p}\mathbf{S}^{(p)}\in\mathbb{R}^{I\times c_s}$ after each block, so that all GPUs hold identical K and V for the next block. The same partition applies at the atom level. The dense atom pair track is striped as $\mathbf{P}^{(p)}\in\mathbb{R}^{L_p\times L\times c_\text{atompair}}$, reducing its per-GPU storage from $\mathcal{O}(L^{2})$ to $\mathcal{O}(L^{2}/P)$. The sparse kNN sequence-local structure-local attention is then applied per-shard: each GPU gathers, from its stripe of $\mathbf{P}^{(p)}$, the $k$ neighbour entries for each of its $L_p$ query atoms, yielding a local bias $\mathbf{P}^{(p)}_\text{sparse}\in\mathbb{R}^{L_p\times k\times c_\text{atompair}}$ at attention-computation cost $\mathcal{O}(L\,k/P)$. At diffusion-step boundaries, rank~0 broadcasts the noised coordinates, sampled Gaussian noise, and denoised prediction so that all ranks share a single stochastic trajectory. Importantly, per-GPU pair-track memory is $\mathcal{O}(I^{2}/P)$ for $\mathbf{Z}$ and $\mathcal{O}(L^{2}/P)$ for $\mathbf{P}$. A more detailed description of this process is provided in the Appendix~\ref{app:1d-impl}.


\subsection{2D grid context parallelism}
\label{sec:2d}

Our second scheme adopts the Fold-CP framework~\cite{lin_fold-cp_2026} and specialises it to RFD3 (\cref{fig:large-monomer-quality}a). We arrange $P$ GPUs on a $\sqrt{P}\times\sqrt{P}$ grid and tile the pair tensor into local quadrants $\mathbf{Z}^{(r,c)}\in\mathbb{R}^{I_r\times I_c\times c_z}$ at grid position $(r,c)$, with $I_r=I/\sqrt{P}$. Queries are sharded along the row axis and replicated along the column axis; keys and values are sharded along the column axis and circulated by $\sqrt{P}$ ring shifts, with an initial $(r,c)\leftrightarrow(c,r)$ transpose to align them with the query rows. At each ring step, a GPU computes attention between its resident $\mathbf{Q}$ rows and the currently visited K/V shard, using the local bias quadrant, and merges the partial output into a running total via an online softmax. We refer the reader to Fold-CP's \S2--3 and Figure~2 for the full ring-attention derivation, the Cannon-style shift patterns for triangular updates, and the per-module complexity table. All tensors are represented as PyTorch DTensors; parameters are replicated across the grid and verified via a runtime check that every trainable tensor is a DTensor.

Asymptotically, per-device pair-track memory is $\mathcal{O}(I^{2}/P)$  (the same as the 1D scheme) while K/V are ring-rotated rather than replicated, so each device only ever holds an $\mathcal{O}(I/\sqrt{P})$ slab of K/V at a time. The $\sqrt{P}$ ring shifts per attention block are overlapped with local computation.

Relative to Fold-CP, the RFD3 adaptation mainly concerns the sampling loop and the atom-level path: the ring primitive is invoked inside every recycling iteration of every denoising step, with rank-$0$ broadcasts at step boundaries that mirror the 1D scheme, and the sparse atom attention requires a distributed kNN that avoids materialising any $[L,L]$ distance tensor. Concrete descriptions of these adaptations (the distributed kNN, the boundary communicators, and the DTensor parameter distribution) are deferred to Appendix~\ref{app:cp2d-impl}. Importantly, this 2D grid scheme applies only when the available GPU count $P$ is a perfect square, so that devices can be arranged as a $\sqrt{P}\times\sqrt{P}$ mesh.


\section{Results}

\subsection{Design quality and symmetry}
\label{sec:results-symmetry}

\begin{figure*}[!t]
  \centering
  \includegraphics[width=\textwidth]{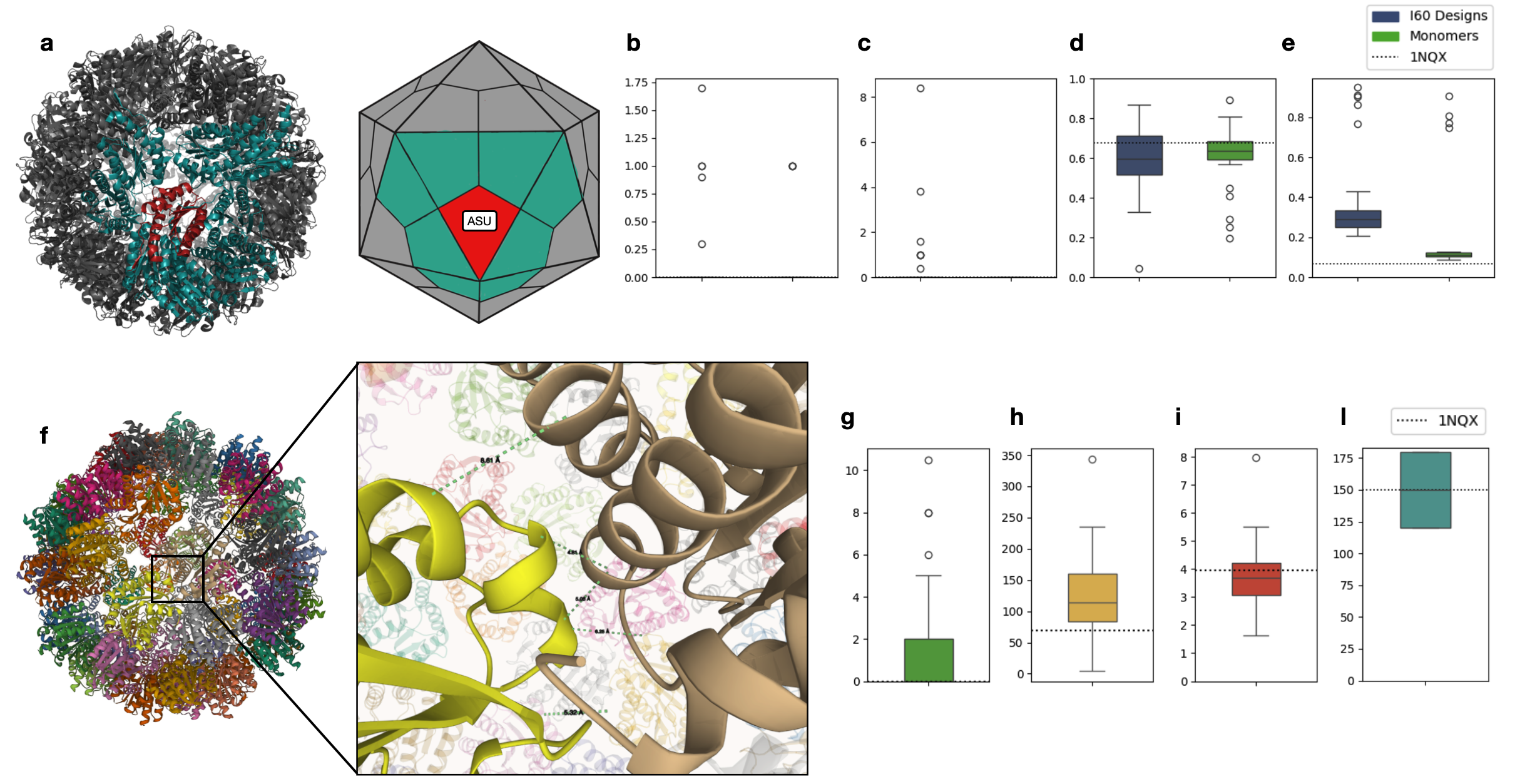}
  \caption{\textbf{Designing icosahedral nanoparticles with Design-CP.} \textbf{a}, Depiction of the ASU (red) and its eight nearest neighbours in the icosahedral assembly (cyan), shown schematically (left) and annotated on a naturally occurring icosahedral protein nanoparticle: Lumazine Synthase (right, PDB: \href{https://www.rcsb.org/structure/1NQX}{1NQX}), which has been used as a scaffold for vaccine development. \textbf{b--e}, Per-chain comparison of Design-CP-generated icosahedral designs (blue, $n=40$, $60$ chains $\times 210$ residues) against original single-GPU RFD3 monomers of length $210$ (green, $n=40$): \textbf{b} average chain breaks, \textbf{c} average backbone clashes, \textbf{d} non-loop fraction, \textbf{e} max CA deviation; the dotted line gives the corresponding value for Lumazine Synthase (1NQX). \textbf{f}, Visualisation of a selected icosahedral design (left) with a zoomed view of the interaction between two ASUs and representative inter-subunit C$\alpha$--C$\alpha$ distances (right). \textbf{g--l}, Symmetry-aware interface metrics for the same $40$ Design-CP-generated icosahedral designs, with the 1NQX reference overlaid: \textbf{g} ASU clashes, \textbf{h} mean contacts per interface, \textbf{i} minimum inter-chain distance, \textbf{l} number of interfaces with contacts. Full metric definitions in Appendix~\ref{app:designability}.}
  \label{fig:sym-neighbourhood-quality}
\end{figure*}

Context parallelism can be applied as an \emph{inference-only} modification: it changes how intermediate activations are partitioned and communicated, but does not alter the learned parameters of RFD3. This makes it immediately applicable to pretrained checkpoints, but also means that when CP is used to exceed RFD3's native crop limits, the model is sampled outside the regime it was trained on (384 tokens and 5000 atoms, according to \S1.6 of the supplementary material of~\citet{butcher_novo_2025}). In practice, we find that this distribution shift can degrade sample quality when the target displays limited symmetry.

\cref{fig:large-monomer-quality}b qualitatively illustrates this effect: when the number of atoms/tokens modelled is much higher than RFD3 saw during training (in this example, designing a single monomer with 10800 amino acids and 2D sharding), we observe visibly unnatural designs. By contrast, when introducing constraints through symmetry, the resulting assemblies appear visibly more protein-like, even while keeping the system size constant. For icosahedral targets, this observation is supported quantitatively in \cref{sec:results-quality}. We also assess that this apparent increase in sample quality is not an effect due to chain length by designing and visually checking a series of asymmetric designs (Appendix \cref{app:asymmetric_multi_chains}) having the same number of chains (60) and same amino acids per chain (180) as the icosahedron shown in \cref{fig:large-monomer-quality}b.

We attribute this effect to the symmetry sampling mechanism described in \cref{sec:prelims}: at each symmetrised denoising step the network performs a full forward pass over all atoms of the assembly, but only the asymmetric subunit (ASU) of its prediction is retained, and the remaining subunits are overwritten by deterministic group-operation copies of that ASU. The sampler therefore varies the coordinates of a single ASU rather than those of the full assembly, which shrinks the design space and couples distant regions of the assembly by forcing every subunit to share the same ASU backbone.

This motivates the usability of CP for the design of highly symmetric protein assemblies such as octahedral and icosahedral capsids and cages. CP allows the model to consider (and remain self-consistent with respect to) the full set of residue/atom interactions in the assembly, while the symmetry prior ensures that the number of \emph{free} coordinates the sampler must generate is that of a single ASU. We next focus our studies on icosahedral and octahedral protein nanoparticle design.

\subsection{Designing icosahedral nanoparticles}\label{sec:results-quality}

A standard \emph{de novo} design pipeline often couples a structure generator (RFDiffusion/RFD3), a sequence designer (e.g., ProteinMPNN), and an external structure predictor for refolding-based validation. For very large multimeric assemblies, however, prediction-based oracles become both more expensive and less reliable: AlphaFold-Multimer accuracy, for example, has been found to degrade with chain count~\cite{bryant_predicting_2022}. We therefore assess design quality primarily through \emph{in silico} structural sanity checks computed directly from the generated all-atom coordinates, and we organise the assessment into two questions: (i) does sampling at this scale via Design-CP preserve the per-chain backbone quality of the original RFD3 model, and (ii) does the symmetrised assembly realise icosahedral interfaces consistent with a functional natural baseline.

We generated $n=40$ icosahedral nanoparticles ($210$ residues per chain, $60$ chains, $12{,}600$ residues per assembly) with the 2D sharding scheme on a $2\times 2$ grid of HG200 GPUs ($95$\,GB each) at batch size one. As a paired control for question (i), we additionally sampled $n=40$ single-chain $210$-residue monomers with the standard single-GPU RFD3. Symmetry-aware metrics for question (ii) are computed between the ASU and its eight nearest icosahedral neighbours (\cref{fig:sym-neighbourhood-quality}a). The corresponding Lumazine Synthase (PDB:~\href{https://www.rcsb.org/structure/1NQX}{1NQX}) values are overlaid as a dotted reference line in every panel. This particular icosahedral nanoparticle was selected as an example of a naturally occurring nanoparticle that has previously been employed as a vaccine scaffold ~\cite{ladenstein_second_2020, joseph_lumazine_2025}. Full metric definitions and additional supporting plots are reported in Appendix~\ref{app:designability} and~\ref{app:vanilla-rfd3-vs-designcp}.

\begin{table*}[!t]
  \centering
  \small
  \caption{\textbf{Scaling of Design-CP under icosahedral symmetry.} For each GPU count $P$, we report the maximum ASU length before out-of-memory and the wall-clock inference time \emph{at that maximum ASU length}, for the 1D and 2D sharding schemes. \emph{Capacity ratio} is defined as $L_{\max}^{\mathrm{2D}}/L_{\max}^{\mathrm{1D}}$ and \emph{speedup} as $t_{\mathrm{1D}}/t_{\mathrm{2D}}$, so that values greater than unity indicate that the 2D scheme outperforms the 1D scheme. All measurements use HG200 GPUs (95\,GB each); a dash denotes a configuration that cannot be measured.}
  \label{tab:scaling}
  \begin{tabular}{rcccccc}
    \toprule
    & \multicolumn{3}{c}{\textbf{Max ASU length before OOM} (residues)} & \multicolumn{3}{c}{\textbf{Wall-clock time at max ASU} (s)} \\
    \cmidrule(lr){2-4} \cmidrule(lr){5-7}
    $P$ & \textbf{1D} & \textbf{2D} & \textbf{Capacity ratio} & \textbf{1D} & \textbf{2D} & \textbf{Speedup} \\
    \midrule
    1  & 58  & 58  & $1.00\times$ & 1023 & 1025 & $1.00\times$ \\
    2  & 102 & --  & --           & 2340 & --   & --           \\
    3  & 120 & --  & --           & 2943 & --   & --           \\
    4  & 141 & 137 & $0.97\times$ & 4321 & 2230 & $1.94\times$ \\
    5  & 160 & --  & --           & 3845 & --   & --           \\
    6  & 173 & --  & --           & 3662 & --   & --           \\
    7  & 186 & --  & --           & 4802 & --   & --           \\
    8  & 196 & --  & --           & 4689 & --   & --           \\
    9  & 201 & 201 & $1.00\times$ & 7247 & 3253 & $2.23\times$ \\
    16 & 219 & 237 & $1.08\times$ & 7030 & 3785 & $1.86\times$ \\
    \bottomrule
  \end{tabular}
\end{table*}

\paragraph{Per-chain backbone quality broadly matches the original RFD3.} On the backbone-sanity indicators (\cref{fig:sym-neighbourhood-quality}b--d), the two populations have broadly similar distributions: the average number of chain breaks per chain and the average count of backbone heavy-atom clashes per chain both have medians at zero. The ASUs extracted by the nanoparticles designed with Design-CP show, however, more outliers than the designed monomers, especially for the number of backbone clashes. The fraction of residues assigned to a helix or $\beta$-strand secondary-structure element is closely matched between them and to the 1NQX reference (medians of $\approx 0.6$ for icosahedral nanoparticles' ASUs vs.\ $\approx 0.65$ for monomers, against $\approx 0.68$ for 1NQX). Another notable gap is in the maximum per-chain deviation of consecutive C$\alpha$--C$\alpha$ bond lengths from the canonical $3.8$\,\AA{} value (\cref{fig:sym-neighbourhood-quality}e): monomers have a distribution concentrated around $\approx 0.15$\,\AA{}, whereas icosahedral nanoparticles' ASUs chains show a broader bulk around $\approx 0.30$\,\AA{}. We partially attribute the difference in some of the analysed metrics between Design-CP ASUs and single-GPU RFD3 monomers to the geometric strain of having to remain self-consistent with symmetry mates and their inter-subunit interfaces inside a $12{,}600$-residue joint context, rather than to a degradation introduced by the context-parallel sampler itself. The icosahedral design problem is strictly harder per chain than designing monomers of the same length. The full eight-panel comparison and a more detailed discussion are in Appendix~\ref{app:vanilla-rfd3-vs-designcp}.

\paragraph{Symmetric interfaces track a functional natural assembly.} Symmetry-aware metrics (\cref{fig:sym-neighbourhood-quality}g--l) generally follow the 1NQX baseline, but not every sample produces a sterically clean assembly. The number of inter-subunit clashes within the ASU's neighbourhood (\cref{fig:sym-neighbourhood-quality}g) is non-zero for a substantial fraction of designs, despite the mean being around zero. Concordantly, the minimum C$\alpha$--C$\alpha$ distance between distinct chains (\cref{fig:sym-neighbourhood-quality}i) is centred near the 1NQX reference at $\approx 4$\,\AA{}, but the lower tail of the distribution descends toward and below the $3.5$\,\AA{} steric-overlap threshold defined in Appendix~\ref{app:designability}, indicating that a fraction of designs realise inter-chain contacts inside the steric-overlap regime. The average number of C$\alpha$--C$\alpha$ contacts per inter-subunit interface centres around $\approx 125$  (\cref{fig:sym-neighbourhood-quality}h), slightly above 1NQX rather than collapsing toward zero, and the average number of contacts per interface resembles that of 1NQX (\cref{fig:sym-neighbourhood-quality}l). Combined with the local visual evidence from one selected sample in \cref{fig:sym-neighbourhood-quality}f, where ASU--ASU contacts settle into well-formed packing geometries with C$\alpha$--C$\alpha$ distances in the expected $\approx 4$--$10$\,\AA{} band, this indicates that the symmetrised denoising trajectory can recover physically reasonable inter-subunit interfaces rather than independently designing non-interacting chains.

Together, these results show that Design-CP scales RFD3 sampling to large icosahedral assemblies without retraining or fine-tuning, broadly preserving per-chain backbone quality on par with vanilla RFD3 monomers and producing symmetry-consistent interfaces comparable to a functional natural icosahedral particle.

\begin{figure*}[!t]
  \centering
  \includegraphics[width=\textwidth]{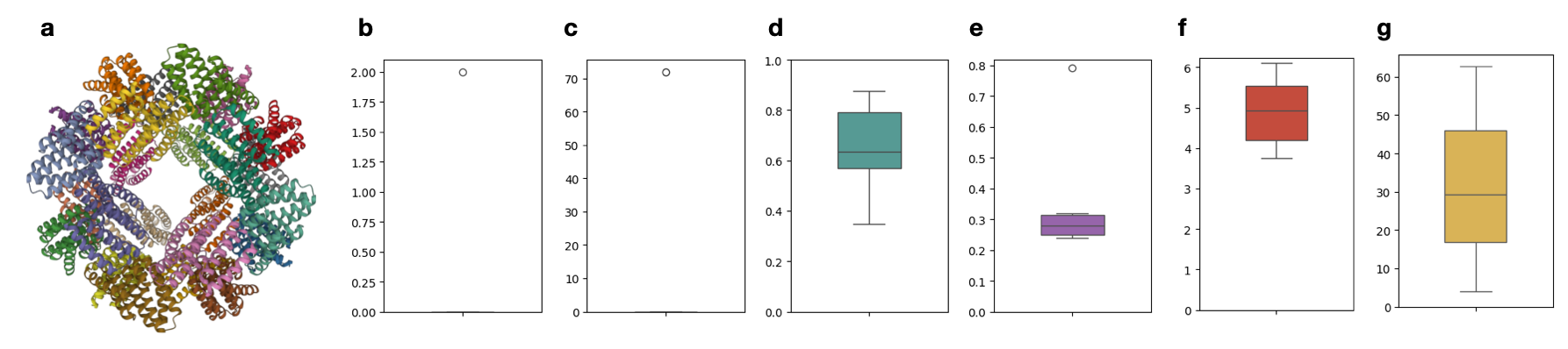}
  \caption{\textbf{Designing octahedral nanoparticles on workstation-grade GPUs.} \textbf{a}, A representative Design-CP octahedral assembly with $176$ residues per chain ($24$ chains, $4{,}224$ residues total) generated on $16$ NVIDIA RTX~A4000 GPUs ($16$\,GB per device). \textbf{b--e}, Backbone-sanity metrics over $n=12$ such designs: \textbf{b} chain breaks, \textbf{c} backbone clashes, \textbf{d} non-loop fraction, \textbf{e} max CA deviation. \textbf{f--g}, Symmetry-interface metrics: \textbf{f} minimum inter-chain distance, \textbf{g} mean contacts per interface. Metric definitions in Appendix~\ref{app:designability}; the remaining secondary-structure and contact metrics are reported in Appendix~\ref{app:O24_scaling_and_metrics}.}
  \label{fig:octahedral-mainfig}
\end{figure*}

\subsection{Scaling: max ASU size and inference time vs. number of GPUs}
\label{sec:results-scaling}

To validate the extent of applicability of our proposed CP techniques, we performed a scaling analysis on the task of designing proteins with icosahedral point-group symmetry. Specifically, we target the icosahedral symmetry and sweep the amino-acid length of the asymmetric subunit (ASU) until inference runs out of memory (OOM) for a fixed number of GPUs $P$. \Cref{tab:scaling} shows these values and the inference time for that specific run right before reaching OOM error. Unless explicitly mentioned, all experiments in this subsection use HG200 GPUs with 95\,GB memory and a batch size of one.

\paragraph{Maximum ASU size.}
With either sharding scheme, the dominant quadratic pair tracks are distributed across the device mesh, so the memory ceiling is set primarily by the \emph{aggregate} available memory rather than by a single-device limit.  We believe this is the reason why we observe this value of max ASU length before OOM to be similar, independently of the sharding scheme used.  Empirically, the maximum feasible ASU length increases with $P$ with the expected square-root trend predicted in \Cref{sec:1d,sec:2d}: as $P$ grows linearly while all the GPUs are maximally used, each device stores a quadratically smaller proportion of the growing pair tensors. 

\paragraph{Inference time.}
While the two schemes reach comparable memory ceilings at a given $P$, their wall-clock scaling differs. The 1D scheme performs a single \textsc{AllGather} of the single-track activations after each attention block (\cref{sec:1d}), which introduces a latency cost that grows with cluster size and impacts runtime. The 2D scheme replaces global gathers with $\sqrt{P}$-step ring communication over smaller shards and overlaps these shifts with local computation (\cref{sec:2d}), yielding better time scaling in practice. This effect is reflected in \Cref{tab:scaling}, where 2D CP achieves lower per-step inference time at the same $P$, often finishing inference in around half the time with respect to the 1D inference.


\subsection{Designing octahedral nanoparticles on small GPUs}
\label{sec:results-democratization}

Octahedral protein cages are a promising but underexplored therapeutic scaffold class: their rigid, multivalent geometry is suited to higher-order receptor engagement, and their interior volume can encapsulate biologic cargoes. Despite this potential, only a few \emph{de novo} octahedral cages have been characterised in therapy-relevant cell-based functional assays~\cite{divine_designed_2021, yang_computational_2024}, and as noted in \cref{sec:introduction}, the memory cost of jointly sampling such large symmetric assemblies has been a practical barrier to broadening this design space.

To test whether Design-CP lifts this barrier on commodity hardware, we repeated the scaling protocol of \cref{sec:results-scaling} for octahedral targets, this time using a small cluster of workstation-grade NVIDIA RTX~A4000 GPUs ($16$\,GB per device, ${\sim}6\times$ less per-device memory than the HG200 GPUs used so far) and we report the results in \cref{app:O24_scaling_and_metrics}. In practice, the 2D sharding scheme allowed us to sample full octahedral assemblies up to $178$ residues per chain (with $24$ chains, this means modelling $4{,}272$ residues total) on $16$ A4000 GPUs (\cref{fig:octahedral-mainfig}a), a per-chain size comparable to existing \emph{de novo} octahedral nanoparticles~\cite{king_computational_2012}; we then evaluated $n=12$ such designs against the same families of \emph{in silico} metrics used for the icosahedral targets in \cref{sec:results-quality}. Additional metrics and a discussion of the scaling sweep on this hardware are deferred to Appendix~\ref{app:O24_scaling_and_metrics}.

\paragraph{Backbone sanity tracks the icosahedral baseline.} The backbone quality indicators (\cref{fig:octahedral-mainfig}b--e) are healthy and consistent with the icosahedral designs population: the per-chain count of chain breaks concentrates at zero with a single outlier at $2$, the count of backbone heavy-atom clashes is essentially zero across all designs (one outlier near $70$), the fraction of residues assigned to a helix or strand element sits at a median of $\approx 0.65$, and the maximum per-chain deviation of consecutive C$\alpha$--C$\alpha$ bond lengths from the canonical $3.8$\,\AA{} value clusters tightly around $\approx 0.3$\,\AA{}, well below the $0.75$\,\AA{} chain-break threshold. This last distribution is in fact noticeably tighter than for the icosahedral chains, and might suggest that the lower oligomeric state ($24$ vs.\ $60$ subunits) imposes less geometric strain per chain.

\paragraph{Symmetric interfaces are well-formed.} The two main symmetry-interface metrics shown in \cref{fig:octahedral-mainfig}f--g are likewise healthy. The minimum C$\alpha$--C$\alpha$ distance between distinct chains sits in the $\approx 4$--$10$\,\AA{} band, consistent with proper inter-subunit packing without steric overlap, and the average number of C$\alpha$--C$\alpha$ contacts per inter-subunit interface centres around $\approx 30$ with a positive tail beyond $60$. An example of octahedral design is provided in \cref{fig:octahedral-mainfig}a, where the chains organise into a closed, octahedrally symmetric cage. This confirms that the symmetrised trajectory realises the intended point-group geometry with favourable in-silico metrics on commodity GPUs.

Overall, these results indicate that large-assembly protein design can be made feasible on modest, widely available GPU hardware, lowering the barrier to entry for groups without access to large-memory accelerators.

\section{Conclusion}
\label{sec:conclusion}

We introduced \emph{Design-CP}, two context-parallel inference strategies for RFDiffusion~3 that shard the quadratic token- and atom-pair representations across multiple GPUs while preserving the model's architecture and weights. We characterised the memory ceilings and strong-scaling behaviour of a lightweight 1D row-sharding scheme and a 2D grid scheme (which requires a perfect-square GPU count $P$ to form a $\sqrt{P}\times\sqrt{P}$ mesh), and found that 2D sharding achieves better wall-clock scaling in practice by overlapping ring communication with computation.

Beyond RFD3, the underlying approach is model-agnostic: any protein design model whose inference is dominated by self-attention and other $\mathcal{O}(n^2)$ pairwise activations can, in principle, adopt the same sharding patterns. This suggests that context parallelism is broadly applicable across modern design pipelines, which largely build on Transformer-style attention mechanisms.

We further showed that strong point-group symmetry constraints can make CP usable \emph{out of the box} for end-to-end all-atom design of large protein nanoparticles without retraining or fine-tuning. Prior work on context-parallelism for structure prediction (e.g., Fold-CP~\cite{lin_fold-cp_2026}) indicates that scaling inference to very large complexes can be limited by out-of-distribution generalisation once inputs exceed the model's native training regime. Our results refine this picture: for highly symmetric assemblies, symmetrised sampling collapses the effective degrees of freedom to a single ASU while still modelling the correct inter-subunit geometry, enabling CP to scale to full icosahedral and octahedral cages while preserving promising \emph{in silico} backbone-sanity and symmetry-interface metrics (\cref{sec:results-quality,sec:results-democratization}).

This changes the design paradigm: whereas existing nanoparticle pipelines often rely on rigid-body docking of pre-computed oligomeric building blocks, Design-CP enables joint denoising of all atoms of all asymmetric subunits and their inter-subunit interactions within a single trajectory. We also demonstrated that the same approach enables \emph{de novo} design of octahedral nanoparticles on a small cluster of workstation-grade GPUs, indicating a practical path towards democratising large-assembly protein design.

A key limitation is that pushing beyond the native training crop can still introduce distribution shift and degrade outputs for highly asymmetric targets. Future work should therefore explore training or fine-tuning design models with longer contexts and/or explicit context-parallelism in the training loop, as well as experimental validation of the designed assemblies to assess how in-silico quality metrics translate to expression, stability, and correct self-assembly \emph{in vitro}.

\section*{Acknowledgements}
L.T. acknowledges support from Ellision Institute of Technology, Oxford Ltd. The authors are grateful to Prof. David Baker and Prof. Frank DiMaio for valuable discussions, and to the Institute of Protein Design at the University of Washington for access to computational resources. The authors acknowledge the use of resources provided by the Isambard-AI National AI Research Resource (AIRR). Isambard-AI is operated by the University of Bristol and is funded by the UK Government’s Department for Science, Innovation and Technology (DSIT) via UK Research and Innovation; and the Science and Technology Facilities Council [ST/AIRR/I-A-I/1023].

\section*{Impact Statement}




This work studies inference-time scaling for all-atom generative protein design models. By distributing the dominant quadratic activations across a multi-GPU mesh, Design-CP makes end-to-end design of large symmetric protein assemblies feasible on hardware ranging from data-centre clusters to workstation-grade GPUs. We see the primary positive impact as democratising access to large-assembly design: lowering the hardware barrier broadens participation beyond well-resourced groups and could accelerate progress in therapeutic delivery systems, vaccine scaffolds, and other biomedical applications of protein nanoparticles. The contributions are at the inference and parallelisation layer and do not introduce new model weights, training data, or predictive capabilities. The same broadening of access does, however, raise dual-use concerns: tools that make symmetric-assembly design easier could, in principle, be misused for harmful protein engineering. We view such risks as best mitigated through standard biosecurity governance, responsible disclosure norms, and careful application review at the point of use.


\bibliography{references}
\bibliographystyle{icml2026}


\appendix
\onecolumn

\section{RFD3 background}
\label{app:rfd3-bg-parent}

This appendix expands on \S\ref{sec:prelims} by recording the architectural choices, hyper-parameters, and inference-loop mechanics of stock RFDiffusion 3 that Design-CP wraps without modification. The intent is to make the rest of the paper readable in isolation: every constant referenced in \S\ref{sec:methods}--\S\ref{sec:results-democratization} is fixed here, with values taken from the configuration files of the open-source codebase.

\subsection{RFDiffusion 3 architecture and denoising loop}
\label{app:rfd3-bg}

\paragraph{Token and atom representations.} As in the main text, RFD3~\cite{butcher_novo_2025} jointly maintains a token-level single track $\mathbf{S}\in\mathbb{R}^{I\times c_s}$ and pair track $\mathbf{Z}\in\mathbb{R}^{I\times I\times c_z}$, together with an atom-level single track $\mathbf{A}\in\mathbb{R}^{L\times c_\text{atom}}$ and dense pair track $\mathbf{P}\in\mathbb{R}^{L\times L\times c_\text{atompair}}$. The default channel widths in the open-source configuration are $c_s=384$, $c_z=128$, $c_\text{atom}=128$, and $c_\text{atompair}=16$. An additional internal channel $c_\text{token}=768$ is used inside the diffusion path for the upcast/downcast cross-attention modules, and a Fourier time embedding of dimension $c_{t,\text{embed}}=256$ encodes the current noise level.

\paragraph{Pairformer blocks.} The five-stage architecture announced in \S\ref{sec:prelims} is built from two recurring micro-architectures. The Pairformer block used inside the token initialiser and the diffusion token encoder is, in this codebase, an \emph{AttentionPairBias} ($16$ heads, optional QK-norm) followed by a Transition module; both stock configurations explicitly disable the AlphaFold-3-style triangle multiplication and triangle attention paths (\texttt{use\_triangle\_attn=false}, \texttt{use\_triangle\_mult=false}). The atom-attention block used in the atom encoder, the atom decoder, and inside the token initialiser's atom embedder is an \emph{AttentionPairBiasDiffusion} attention layer with $4$ heads followed by a transition; the open-source default applies $0.10$ dropout inside the diffusion-side atom blocks. All blocks share a common \emph{conditional} layer-norm path that injects the time embedding and, where applicable, the conditioning state.

\paragraph{Five-stages architecutre.} With those primitives, the stock open-source configuration realises the following pipeline:
\begin{itemize}
\item \textbf{Token initialiser} (one call per trajectory). A token-level single-track is built from the discrete features (residue type, motif tokens, predicted-LDDT, ``non-loopy'' flag), the pair track $\mathbf{Z}$ is initialised from the outer sum of two independent linear projections of $\mathbf{S}$ plus a relative-position-encoding bias, and an atom embedder pre-aggregates $\mathbf{A}$ into a starting per-token feature. Two \emph{non-triangular} Pairformer blocks (each $16$-head AttentionPairBias + Transition) then refine $(\mathbf{S},\mathbf{Z})$, and the dense atom pair tensor $\mathbf{P}\in\mathbb{R}^{L\times L\times c_\text{atompair}}$ is materialised at this stage.
\item \textbf{Atom encoder} (one call per trajectory). Three blocks of sparse atom attention (each a $4$-head AttentionPairBiasDiffusion + Transition) refine the atom-level latent $\mathbf{A}$ using the kNN budget described below.
\item \textbf{Diffusion token encoder} (one call per recycling iteration). The pair track is augmented along the channel dimension with a noised-coordinate distogram $\mathbf{D}_\text{dist}$ and the self-conditioning distogram $\mathbf{D}_\text{self}\in\mathbb{R}^{B\times I\times I\times n_\text{bins}}$ (with $n_\text{bins}=65$), then passed through two further non-triangular Pairformer blocks. The single track $\mathbf{S}$ receives an AdaLN injection of the Fourier time embedding before each block.
\item \textbf{Diffusion transformer} (one call per recycling iteration). Eighteen sequential AttentionPairBiasDiffusion + ConditionedTransitionBlock blocks ($16$ heads each, $0.10$ dropout) update the token-level latent $\mathbf{A}_I\in\mathbb{R}^{B\times I\times c_\text{token}}$ using the local pair bias derived from $\mathbf{Z}$. This is the stage that dominates per-step compute and that motivates the per-block \textsc{AllGather} in the 1D parallel scheme.
\item \textbf{Atom decoder} (one call per recycling iteration). Three blocks each apply a token-to-atom upcast (cross-attention from atoms onto the current token state, with an $n_\text{split}=3$ chunking of the upcast linear), a $4$-head atom-level self-attention block with $0.10$ dropout, and a token-level downcast that scatter-means atom features back to tokens. The final block emits a per-atom coordinate update which is unwound through the EDM update of \S\ref{app:edm-loop}.
\end{itemize}
Within each denoising step, stages~1--2 fire once, while stages~3--5 are repeated for $n_\text{recycle}=2$ recycling iterations: the first iteration uses a zero-initialised $\mathbf{D}_\text{self}$, and each subsequent iteration feeds back the distogram computed by bucketising the previous iteration's predicted C$\alpha$ coordinates.

\paragraph{Token- and atom-level features at the boundary.} The token initialiser reads a per-token feature dictionary that, in the open-source configuration, sums to $c_{s,\text{inputs}}=37$ channels before projection: a $32$-dim residue-type embedding, a $3$-dim motif-token-type one-hot, a scalar reference plDDT, and a scalar non-loopy flag. The atom embedder consumes a much wider $402$-dim per-atom feature: $256$-dim character-level atom-name encodings, a $128$-dim element embedding, $3$-dim reference coordinates, and a battery of scalar flags (formal charge, occupancy mask, motif-atom-with-fixed-coord, motif-atom-unindexed, has-zero-occupancy) plus the conditioning channels (per-atom RASA, hydrogen-bond donor/acceptor activity, atom-level hotspot). The relative-position-encoding bias added to $\mathbf{Z}_\text{init}$ is built from one-hot encodings of residue offset (range $\pm32$), token offset (same range), chain-hop separation (range $\pm 2$), and a same-entity boolean.

\paragraph{Atom-level kNN attention budget.} The $\mathcal{O}(L\,k)$ atom-level attention announced in \S\ref{sec:prelims} draws its $k$ neighbours from a structured budget: a fixed number $n_\text{seq}$ of per-residue sequence-local neighbours (default $n_\text{seq}=2$, namely the query atom plus its immediate flanking residues' atoms), and the spatially closest atoms beyond those, until the per-query budget $k=n_\text{attn-keys}$ (default $k=128$) is filled. The kNN indices are computed once per denoising step from the current noised coordinates $\mathbf{X}^{(t)}$ via a single $\mathbf{cdist}$ call and reused across all recycling iterations of that step. When the input is a multi-chain assembly with more than three chains, the budget is split into an intra-chain quota of $k - \max(32, k/4)$ neighbours and an inter-chain quota of at least $32$ neighbours, ensuring that every query atom always retains at least $32$ context atoms outside its own chain; this is the mechanism by which the dense $[L,L]$ distogram gets replaced by a sparse $[L,k]$ slice without losing inter-chain interactions.

\paragraph{Chunked pairwise embedder.} The dense $[L,L,c_\text{atompair}]$ allocation that the standard token initialiser performs becomes prohibitive on the assemblies we target. The optional low-memory mode that the main text mentions in passing replaces this dense allocation by a coordinate-dependent on-the-fly construction: every linear projection of $\mathbf{P}$ that depends only on per-atom features (reference coordinates, element, charge, residue-bond graph) is precomputed and \emph{cached} once at tokenisation, while the coordinate-dependent components -- the inverse-distance and same-residue masks that vary with $\mathbf{X}^{(t)}$ -- are recomputed at the kNN indices at the start of every atom-attention block. The block therefore consumes a $[L,k,c_\text{atompair}]$ slice rather than a $[L,L,c_\text{atompair}]$ tensor, at the cost of recomputing the coordinate-dependent embeddings $n_\text{block}$ times per step. This path is gated by the environment variable \texttt{RFD3\_LOW\_MEMORY\_MODE}; both Design-CP schemes auto-enable it whenever \texttt{RFD3\_ATTENTION\_PARALLEL} is set, because the row/quadrant striping reduces $\mathcal{O}(L^2)$ to $\mathcal{O}(L^2/P)$ but only the chunked path pushes the per-rank atom-pair memory all the way down to $\mathcal{O}(Lk/P)$.

\paragraph{EDM denoising loop and Karras parameters.}
\label{app:edm-loop}
Structure generation follows the EDM framework~\cite{karras_elucidating_2022} with the Algorithm-18 schedule from the AlphaFold-3 supplement. The default trajectory is $T=200$ denoising steps with $t\in[0,1]$ linearly spaced, and the per-step noise level is
\begin{equation}
\hat{t} \;=\; \sigma_\text{data}\bigl(s_\text{max}^{1/p} + t\,(s_\text{min}^{1/p}-s_\text{max}^{1/p})\bigr)^{p},
\label{eq:edm-schedule}
\end{equation}
with stock defaults $\sigma_\text{data}=16$, $s_\text{min}=4\!\times\!10^{-4}$, $s_\text{max}=160$, and $p=7$. At each step the sampler optionally injects a Karras-style stochastic noise augmentation of magnitude $\gamma_0=0.6$ when $\hat{t}>\gamma_\text{min}=1.0$ (otherwise $\gamma=0$), perturbs the current state by $\boldsymbol{\epsilon}\sim\mathcal{N}(0,\sigma_\text{noise}^{2}I)$ with $\sigma_\text{noise}=1.003$, calls the diffusion module to obtain the clean prediction $\hat{\mathbf{X}}_0$, and updates
\begin{equation}
\mathbf{X}^{(t+1)} \;=\; \mathbf{X}^{(t)}_\text{noisy} + s\,(c_t-\hat{t})\,\frac{\mathbf{X}^{(t)}_\text{noisy}-\hat{\mathbf{X}}_0}{\hat{t}},
\label{eq:edm-step}
\end{equation}
with step scale $s=1.5$. The recycling loop sits inside this update: between successive denoising steps the model holds the predicted distogram $\mathbf{D}_\text{self}$ obtained by bucketising the predicted C$\alpha$ coordinates (uniform $65$-bin distance grid over $[2,22]\,\textup{\AA}$) and consumes it as the self-conditioning input of the next iteration. The bucketising routine and the self-conditioning channel are unchanged from stock RFD3 and are reused as-is by both Design-CP schemes.

\subsection{Symmetry in RFD3 design}
\label{app:symmetry}

\paragraph{Symmetric inference loop.} The stock RFD3 inference path supports cyclic ($C_n$), dihedral ($D_n$), and arbitrary user-supplied (\texttt{input\_defined}) point groups: the first two are produced analytically from the order $n$ on the fly, while the third is loaded from a user-provided frames file and validated against the ASU at run-time. Each point group is materialised as a list of $G$ rotation matrices $\{R_g\}_{g=1}^{G}\in\mathrm{SO}(3)$ paired with zero translations; for $C_n$, $G=n$ rotations about a common axis are evenly spaced by $2\pi/n$, and for $D_n$, $2n$ rotations combine the cyclic axis with $n$ orthogonal $C_2$ axes. A \texttt{SymmetryConfig} dataclass groups the chosen identifier together with two additional handles -- \texttt{is\_unsym\_motif}, a comma-separated list of contig or ligand identifiers (DNA strands, small-molecule cofactors) that should not be replicated by the symmetry operation, and \texttt{is\_symmetric\_motif}, a Boolean controlling whether the supplied input is already symmetric or must itself be replicated -- and is the single entry point used by both Design-CP schemes.

\paragraph{ASU-based input construction.} Given an ASU atom array, the inference engine first appends the symmetry metadata (a per-atom subunit index and the current frame stack), then walks through the frame list and produces $G$ copies of the ASU by applying each $R_g$ to the ASU coordinates; unsymmetrised motifs (DNA, ligands, any contig matched by \texttt{is\_unsym\_motif}) are excised before the per-frame replication and re-appended at the end of the resulting atom array, which keeps them out of the symmetric coordinate pool while preserving their indexing inside the model. When \texttt{is\_symmetric\_motif=True}, the frames inferred analytically from the symmetry identifier are first reconciled with the empirical frames recovered from the input by aligning corresponding chains via SVD; this provides a sanity check that the user-supplied symmetric input actually obeys the requested point group, and produces the per-frame translations that the analytic axis-aligned frames lack.

\paragraph{Per-step symmetrisation procedure.} At each denoising step within the symmetrised portion of the trajectory (controlled by a \texttt{sym\_step\_frac} parameter, default $0.9$, covering the first $90\%$ of steps): (i)~the \emph{full} (unsymmetrised) noised coordinates $\mathbf{X}^{(t)}$ are passed to the network, which predicts clean coordinates $\hat{\mathbf{X}}_0$ for all chains; (ii)~the predicted coordinates are centred by subtracting the centroid of the non-fixed atoms (atoms tagged by the motif/ligand-exclusion masks above are excluded from the centroid), the ASU slice of $\hat{\mathbf{X}}_0$ is extracted, and the non-ASU chains are overwritten by $R_g\,\hat{\mathbf{X}}_{0,\text{ASU}}$ for $g=2,\dots,G$; (iii)~this symmetrised prediction is used in the EDM update of Eq.~(\ref{eq:edm-step}). The noise injected at each step is \emph{not} symmetrised, so the noised coordinates seen by the network are not exactly symmetric; only the predicted output is forced to be exactly symmetric at each symmetrised step. For the final $10\%$ of steps, no symmetry is enforced, allowing the model to relax any residual strain at the inter-subunit interfaces before output.

\paragraph{Composition with context parallelism.} The procedure above is applied independently on every rank: because rank~$0$ broadcasts the noised coordinates, the sampled Gaussian noise, and the network's clean prediction at every diffusion step (Appendix~\ref{app:nondet} for the 1D scheme, Appendix~\ref{app:2d-loop} for the 2D scheme), the inputs to step (ii) above are bit-identical across ranks, and the deterministic ASU extraction and frame application reproduce the same symmetrised prediction on every device. No additional communication is needed for symmetric design beyond the ones that the parallel schemes already perform. This is the operational sense in which the symmetrisation procedure is \emph{orthogonal} to context parallelism, and it is the reason a single set of pretrained weights designs both the unsymmetrised baselines of \S\ref{sec:results-symmetry} and the icosahedral and octahedral assemblies of \S\ref{sec:results-quality}--\S\ref{sec:results-democratization}.

\section{Design-CP 1D row-sharding implementation details}
\label{app:impl-parent}
\label{app:1d-impl}

This appendix collects the engineering details that make the 1D scheme of \S\ref{sec:1d} work in practice: the chunk-distribution algorithm, the per-collective communication volume, the determinism guard required at the atom$\to$token boundary, the per-component memory optimisations layered on top of row striping, and a few smaller bookkeeping items (class structure, environment variables, checkpoint compatibility). The 2D-specific machinery that adapts the Fold-CP framework to RFD3 is collected separately in Appendix~\ref{app:cp2d-impl}.

\subsection{Chunk distribution algorithm}
\label{app:chunk}

Tokens (or atoms) are distributed across $P$ GPUs using floor division with the remainder assigned to the early ranks:
\begin{equation}
I_p =
\begin{cases}
\lfloor I/P \rfloor + 1 & \text{if } p < I \bmod P, \\
\lfloor I/P \rfloor     & \text{otherwise},
\end{cases}
\quad
\text{start}_p = p \cdot \lfloor I/P \rfloor + \min(p,\, I \bmod P).
\end{equation}
Chunk sizes therefore differ by at most one across GPUs, bounding the load imbalance per attention block to a single row regardless of $P$. The \textsc{AllGather} operations handle uneven chunks by padding each GPU's local tensor to the per-rank maximum before the collective and trimming the concatenated result back to the exact total size $I$. The same algorithm is reused at the atom level with $L$ replacing $I$.

\subsection{Communication volume analysis}
\label{app:communication}

\Cref{tab:comm-1d} details the collective communication operations executed during each recycling iteration of the 1D scheme. All sizes assume \texttt{bfloat16} precision (2~bytes per element). The total per-recycle volume per rank is dominated by the $18$ \textsc{AllGather}($\mathbf{A}$) calls from the diffusion transformer; each rank's message size is set by $I\cdot c_\text{token}$ and is independent of $P$. The single \textsc{Broadcast}($\mathbf{A}_I$) after \texttt{process\_a} is a correctness requirement explained in \S\ref{app:nondet}. Per-rank message size is independent of $P$, but the latency component of NCCL \textsc{AllGather} grows with $P$, so the per-block communication time still increases with the device count even when each rank's payload is held fixed; this latency-bound term is the operative constant behind the 1D vs.\ 2D strong-scaling gap reported in \S\ref{sec:results-scaling}.

\begin{center}
\small
\begin{tabular}{@{}lrll@{}}
\toprule
\textbf{Operation} & \textbf{Count / recycle} & \textbf{Tensor shape} & \textbf{Source module} \\
\midrule
\textsc{AllGather}($\mathbf{S}$)
  & 2 & $[I, c_s]$ & Token initializer\textsuperscript{$\dagger$} \\
\textsc{AllGather}($\mathbf{S}$)
  & 2 & $[I, c_s]$ & Diffusion token encoder \\
\textsc{AllGather}($\mathbf{A}$)
  & 18 & $[I, c_\text{token}]$ & Diffusion transformer \\
\textsc{AllGather}($\mathbf{Q}$)
  & 3 & $[L, c_\text{atom}]$ & Atom encoder \\
\textsc{AllGather}($\mathbf{Q}$)
  & 3 & $[L, c_\text{atom}]$ & Atom decoder \\
\textsc{Broadcast}($\mathbf{A}_I$)
  & 1 & $[B, I, c_\text{token}]$ & After \texttt{process\_a} \\
\textsc{Broadcast}($\hat{\mathbf{X}}, \boldsymbol{\epsilon}, \mathbf{X}_\text{noisy}$)
  & 2--3 & $[B, L, 3]$ & Sampling loop\textsuperscript{$\ddagger$} \\
\textsc{Broadcast}(\texttt{sequence\_logits})
  & 0--1 & $[B, I, n_\text{seq}]$ & Sampling loop\textsuperscript{$\ddagger$} \\
\bottomrule
\end{tabular}
\captionof{table}{Collective communication in 1D parallel inference, counted per recycling iteration. Multiply by $n_\text{recycle}$ (default $2$) to obtain the per-denoising-step cost. \textsuperscript{$\dagger$}The token initialiser fires only once per trajectory, not once per step. \textsuperscript{$\ddagger$}Sampling-loop broadcasts are issued once per denoising step, independently of $n_\text{recycle}$; the optional rotation-augmentation broadcast is what takes the $\hat{\mathbf{X}}/\boldsymbol{\epsilon}/\mathbf{X}_\text{noisy}$ count from $2$ to $3$, and the optional sequence-logits broadcast appears only when sequence design is enabled.}
\label{tab:comm-1d}
\end{center}

\subsection{Non-determinism guard: \texttt{process\_a} broadcast}
\label{app:nondet}

The function \texttt{process\_a} aggregates atom-level features to the token level using \texttt{torch.Tensor.index\_reduce(..., "mean")}. \texttt{index\_reduce} performs atomic floating-point accumulations whose ordering depends on per-device scheduling, and is therefore non-deterministic across GPU devices. In 1D parallel mode, this causes each rank to compute slightly different token-level features $\mathbf{A}_I$ -- the empirical magnitude of these discrepancies is on the order of ${\sim}0.06$ in \texttt{bfloat16}. Because $\mathbf{A}_I$ subsequently serves as keys and values for the cross-attention transformer, the small discrepancies are amplified by the downstream linear projections to ${\sim}0.5$, which corrupts the cross-attention invariant that all ranks must share identical $\mathbf{K}$ and $\mathbf{V}$ for the per-block formulation in Eq.~(\ref{eq:cross_attn}) to produce identical outputs across ranks. The fix is a single \textsc{Broadcast} of $\mathbf{A}_I$ from rank~$0$ to all ranks immediately after \texttt{process\_a} returns, before any downstream operation that depends on $\mathbf{A}_I$ being identical across ranks. The corresponding boundary in the 2D scheme is handled differently and described in Appendix~\ref{app:2d-pooling}.

\subsection{Per-component memory optimisations}
\label{app:memory}

\Cref{tab:memory-1d} summarises the per-GPU memory reductions achieved by each optimisation in the 1D parallel inference pipeline. Three of these (manual SwiGLU decomposition, pre-allocation in place of concatenation, and relative-position-encoding sub-chunking) are non-trivial enough to warrant prose explanations; the remaining three follow directly from row striping and from the chunked pairwise embedder that is auto-enabled together with \texttt{RFD3\_ATTENTION\_PARALLEL} (cf.\ Appendix~\ref{app:implementation}).

\begin{center}
\small
\begin{tabular}{@{}lllll@{}}
\toprule
\textbf{Optimisation} & \textbf{Tensor} & \textbf{Standard shape} & \textbf{Parallel (per GPU)} & \textbf{Factor} \\
\midrule
$\mathbf{Z}$ striping
  & Token pairs
  & $[I, I, c_z]$
  & $[I/P, I, c_z]$
  & $P\times$ \\
$\mathbf{D}_\text{self}$ chunking
  & Self-cond.\ distogram
  & $[B, I, I, n_\text{bins}]$
  & $[B, I/P, I, n_\text{bins}]$
  & $P\times$ \\
$\mathbf{P}$ sparsification
  & Atom pairs
  & $[L, L, c_\text{atompair}]$
  & $[L/P, k, c_\text{atompair}]$
  & $PL/k$ \\
$\mathbf{Z}_\text{aug}$ pre-allocation
  & Concat buffer
  & $3{\times}$ peak from \texttt{cat}
  & $1{\times}$ in-place copy
  & $3\times$ \\
Manual SwiGLU
  & Transition activations
  & All intermediates live
  & Sequential with \texttt{del}
  & see below \\
RPE sub-chunking
  & Rel.\ position encoding
  & $[I_p, I, n_\text{bins}]$
  & 4 sub-chunks
  & $4\times$ \\
\bottomrule
\end{tabular}
\captionof{table}{Memory optimisations in 1D parallel inference. $I$: token count, $L$: atom count, $P$: number of GPUs, $k$: sparse-attention neighbour budget. The asymptotic per-GPU pair-track memory after $\mathbf{P}$ sparsification is $\mathcal{O}(L\cdot k/P)$.}
\label{tab:memory-1d}
\end{center}

\paragraph{Manual SwiGLU decomposition.}
\label{app:swiglu}
The pairwise transition layers use a SwiGLU feed-forward network~\cite{shazeer_glu_2020}:
\begin{equation}
\operatorname{Transition}(\mathbf{Z}) =
  W_3 \bigl(\operatorname{SiLU}(W_1 \operatorname{LN}(\mathbf{Z}))
  \odot W_2 \operatorname{LN}(\mathbf{Z})\bigr),
\end{equation}
where $W_1, W_2 \in \mathbb{R}^{c_z \times 4c_z}$ and $W_3 \in \mathbb{R}^{4c_z \times c_z}$. In the standard residual computation $\mathbf{Z} \leftarrow \mathbf{Z} + \operatorname{Transition}(\mathbf{Z})$, PyTorch retains all intermediate tensors simultaneously, including the $4{\times}$-expanded linear outputs. We manually decompose the computation with explicit deallocation (\Cref{alg:swiglu-decomp}). This is mathematically identical but ensures that at most two $4c_z$-expanded tensors coexist at any point, substantially reducing peak memory. The optimisation is applied only during inference (gated on \texttt{torch.is\_grad\_enabled() == False}); training uses the standard path to preserve compatibility with activation checkpointing.

\begin{algorithm}[tb]
  \caption{Manual SwiGLU decomposition (memory-friendly inference implementation).}
  \label{alg:swiglu-decomp}
  \begin{algorithmic}
    \STATE {\bfseries Input:} residual activations $\mathbf{Z}$; weights $W_1, W_2, W_3$
    \STATE {\bfseries Output:} updated activations $\mathbf{Z}$
    \STATE $\mathbf{N} \gets \operatorname{LayerNorm}(\mathbf{Z})$
    \STATE $\mathbf{A} \gets W_1\mathbf{N}$
    \STATE $\mathbf{B} \gets W_2\mathbf{N}$
    \STATE {\bfseries Free:} $\mathbf{N}$
    \STATE $\mathbf{A} \gets \operatorname{SiLU}(\mathbf{A})$
    \STATE $\mathbf{A} \gets \mathbf{A} \odot \mathbf{B}$
    \STATE {\bfseries Free:} $\mathbf{B}$
    \STATE $\mathbf{A} \gets W_3\mathbf{A}$
    \STATE $\mathbf{Z} \gets \mathbf{Z} + \mathbf{A}$
    \STATE {\bfseries Free:} $\mathbf{A}$
  \end{algorithmic}
\end{algorithm}

\paragraph{Pre-allocation vs concatenation.}
\label{app:prealloc}
The diffusion token encoder constructs an augmented pairwise representation by concatenating three components along the channel dimension:
\begin{equation}
\mathbf{Z}_\text{aug} = \bigl[\mathbf{Z}_\text{init} \;\|\;
  \mathbf{D}_\text{dist} \;\|\; \mathbf{D}_\text{self}\bigr]
  \in \mathbb{R}^{B \times I_p \times I
    \times (c_z + c_z + n_\text{bins})}.
\end{equation}
A naive \texttt{torch.cat} requires allocating the output tensor \emph{in addition to} all three inputs, causing a transient memory spike that more than triples the peak at this stage. The 1D parallel implementation pre-allocates a single tensor of the final shape using \texttt{torch.empty} and writes each component into its designated channel slice in-place, deleting the source tensor immediately after each copy. This avoids the concatenation overhead and reduces the peak memory of the augmentation step from $3\times$ to $1\times$ the size of $\mathbf{Z}_\text{aug}$.

\paragraph{Relative position encoding sub-chunking.}
The relative-position-encoding (RPE) bias requires materialising one-hot tensors of shape $[I_p, I, n_\text{bins}]$, which can be large even after row striping. The 1D parallel implementation processes the RPE in $4$ sub-chunks along the query dimension: each sub-chunk computes its slice of the residue, token, and chain one-hot encodings, applies the linear projection that turns them into a single $c_z$-channel bias, and deletes the intermediates before proceeding to the next sub-chunk. This reduces the peak memory of the RPE computation by $4\times$.

\subsection{Class structure and environment variables}
\label{app:implementation}

The 1D scheme is realised through two parallel classes -- \texttt{ParallelTokenInitializer} (which subclasses \texttt{TokenInitializer}) and \texttt{ParallelDiffusionModule} (which subclasses \texttt{RFD3DiffusionModule}) -- that share identical learned parameters with their serial counterparts but override \texttt{forward()} to operate on row-striped representations. A third class, \texttt{ParallelDiffusionTokenEncoder}, is created via a Python \texttt{\_\_class\_\_} swap inside \texttt{ParallelDiffusionModule.\_\_init\_\_}: the standard \texttt{DiffusionTokenEncoder} instance is reassigned to the parallel subclass without re-loading parameters, which is safe because the parallel variant introduces no new parameters and only replaces the forward pass. As a consequence, checkpoints trained on a single GPU are loaded directly without any conversion or key remapping.

A factory in \texttt{RFD3.\_\_init\_\_} reads the environment at construction time and instantiates the parallel classes whenever the relevant flag is set. The 1D scheme is controlled primarily by three environment variables. \texttt{RFD3\_ATTENTION\_PARALLEL} enables parallel mode and selects the world size; the launcher script sets it to the detected GPU count when \texttt{inference\_parallel=True} is passed on the command line, after auto-relaunching the script under \texttt{torchrun} via \texttt{maybe\_relaunch\_with\_torchrun}. \texttt{RFD3\_LOW\_MEMORY\_MODE} enables the chunked pairwise embedder that avoids materialising the dense $[L,L,c_\text{atompair}]$ atom pair tensor; it is auto-enabled whenever \texttt{RFD3\_ATTENTION\_PARALLEL} is set, because the row-striped path requires sparse $\mathbf{P}$ to fit the largest assemblies that motivate context parallelism. \texttt{RFD3\_NCCL\_TIMEOUT\_SEC} is an optional per-collective NCCL watchdog timeout, defaulting to $1800$\,s -- raised from PyTorch's $600$\,s default to tolerate the serial post-processing that rank~$0$ performs (atom-array cleanup, file I/O) while other ranks have already enqueued the next collective. A fourth optional flag, \texttt{RFD3\_EXTRA\_CHUNKING}, controls a finer sub-chunking inside the chunked pairwise embedder for very large $L$ but is rarely needed in practice.

The setup sequence is: (i) the launcher detects \texttt{inference\_parallel=True} in \texttt{sys.argv} and re-execs the script under \texttt{torchrun} with \texttt{--nproc\_per\_node} equal to the visible GPU count; (ii) child processes call \texttt{setup\_distributed()}, which initialises an NCCL process group with the elevated timeout; (iii) the engine sets \texttt{RFD3\_LOW\_MEMORY\_MODE=1} and \texttt{RFD3\_ATTENTION\_PARALLEL=}\textit{world\_size}; (iv) \texttt{RFD3.\_\_init\_\_} reads those variables and constructs the parallel module tree; (v) the engine broadcasts model parameters from rank~$0$ to all other ranks before inference begins, since the Lightning Fabric \texttt{SingleDeviceStrategy} we use does not implicitly replicate weights.

\section{Design-CP 2D grid implementation details}
\label{app:cp2d-impl}

This appendix expands the RFD3-specific aspects of the 2D scheme of \S\ref{sec:2d}. The Fold-CP framework~\cite{lin_fold-cp_2026} supplies the ring-attention core -- transposition of K/V shards at the start of a ring loop, $\sqrt{P}$ ring shifts with online-softmax merging, and the boundary communicators -- which we reuse unchanged. The contributions documented below concern (i) how that core is plugged into RFD3's denoising loop, (ii) how the atom-level sparse attention is made compatible with 2D sharding, (iii) how the dense pre-pipeline data structures of RFD3 are deferred so that the row/column quadrants can be reconstructed on rank, and (iv) how parameters and gradients are laid out on the device mesh. We will release the accompanying code in a future update to the official RFDiffusion 3 repository (https://github.com/RosettaCommons/foundry/tree/production)

\subsection{Device mesh and DTensor parameter distribution}
\label{app:2d-mesh}

Devices are arranged on a $\sqrt{P}\times\sqrt{P}$ context-parallel mesh, optionally combined with a third data-parallel axis when training; we refer to the two CP axes as $\text{cp}_0$ (the row axis) and $\text{cp}_1$ (the column axis). The mesh and the corresponding process subgroups are produced by a \texttt{DistributedManager} object that the engine instantiates before constructing the model; the perfect-square requirement on $P$ is enforced inside the manager, and the model code only ever consumes the precomputed \texttt{device\_mesh\_subgroups} and \texttt{layout\_subgroups} dictionaries.

At model construction, every \texttt{Linear}, \texttt{LayerNorm}, and RMSNorm-shaped module in the serial RFD3 tree is replaced by a parameter-replicated DTensor wrapper (\texttt{LinearParamsReplicated}, \texttt{LayerNormParamsReplicated}) so that its weights live as PyTorch DTensors with a \texttt{Replicate()} placement on every CP axis. A runtime check, \texttt{validate\_all\_params\_are\_dtensors}, traverses the entire module tree before inference begins and asserts that no trainable tensor has silently escaped the wrapping; this catches any custom layer that constructs parameters outside the standard \texttt{nn.Linear}/\texttt{nn.LayerNorm} paths.

The 2D scheme does not use a \texttt{\_\_class\_\_} swap. Instead, a top-level helper \texttt{create\_rfd3\_distributed} instantiates a small set of CP wrapper modules (\texttt{CPTokenInitializer}, \texttt{CPTokenTransformer}, \texttt{CPDiffusionTokenEncoder}, \texttt{CPDiffusionModule}) and replaces the \texttt{forward} attribute of the corresponding serial submodule with the wrapper's \texttt{forward} method. This composition pattern matches Fold-CP's convention and avoids touching the construction-time logic of the serial classes, so checkpoints trained on a single GPU are loaded unchanged.

\subsection{Q/K/V layout and the ring loop}
\label{app:2d-ring}

At each attention block, queries are sharded along $\text{cp}_0$ and replicated along $\text{cp}_1$, while keys and values are sharded along $\text{cp}_1$ and ring-rotated. The ring loop is opened by a single \texttt{TransposeComm} that swaps K/V between grid positions $(r,c)$ and $(c,r)$, aligning the K/V shards with the resident Q rows. The loop then performs $\sqrt{P}$ ring shifts: at each step a GPU computes attention between its resident Q rows and the currently visited K/V shard, using the local pair-bias quadrant; partial outputs are merged into a running total via the online-softmax kernel \texttt{tiled\_softmax\_attention\_update} reused from Fold-CP. Each ring step issues an \texttt{AttentionPairBiasComm}, which bundles the K/V/$\mathbf{B}$ shift and overlaps the communication with local computation, and a small number of \texttt{One2OneComm} point-to-point exchanges that move auxiliary tensors (token indices, chain identifiers, distogram features) along the same column path as the K/V tiles.

Asymptotically, per-device pair-track memory is $\mathcal{O}(I^{2}/P)$ -- the same as the 1D scheme -- but K/V are ring-rotated rather than replicated, so each device only ever holds an $\mathcal{O}(I/\sqrt{P})$ slab of K/V at a time; the pair tensor itself is therefore stored as a local quadrant $[I_r, I_c, c_z]$ with $I_r=I_c=I/\sqrt{P}$, and is never gathered to its full $[I, I]$ shape on any rank.

\subsection{Distributed kNN for sparse atom attention}
\label{app:2d-knn}

RFD3's atom-level attention selects the $k$ nearest neighbours of every query atom from the current predicted coordinates. A naive implementation computes an $[L, L]$ distance matrix and takes the top-$k$ per row; at the scales where 2D CP is useful, that distance matrix is exactly the object we cannot afford. Our distributed kNN proceeds in three stages. First, the small 1D feature tensors -- token identifiers and chain assignments, each of shape $[L]$ -- are allgathered along $\text{cp}_0$ so every rank has global indexing for downstream masking; these are $\mathcal{O}(L)$ in size and cheap to replicate. Second, each rank computes Euclidean distances between its local atom rows and the atoms currently resident on its column partner, producing a per-quadrant distance block of shape $[L_r, L_c]$ with $L_r=L_c=L/\sqrt{P}$; this is implemented as a chunked \texttt{cdist} (default chunk size $1024$ rows) so that even the per-quadrant block is never materialised in full. Third, a \texttt{ring\_topk} primitive rotates partial top-$k$ candidates along $\text{cp}_1$; at each ring step, the local top-$k$ candidates are merged with incoming candidates from the column neighbour to produce a running global top-$k$ per query atom. After $\sqrt{P}$ ring steps, every row-resident rank holds the global top-$k$ indices for its own query atoms, and no further broadcast is needed because the downstream sparse ring attention consumes the indices in place. The end-to-end computation never materialises an $[L, L]$ tensor on any device.

The indices then feed a sparse ring attention (\texttt{sparse\_ring\_attention\_forward}). At each ring step, the global indices are filtered to the current column block, K/V/$\mathbf{B}$ entries are gathered at those filtered indices, and the resulting $[D, H, L_r, k]$ logits are merged into the running softmax via the same online-softmax kernel used by the dense token-level ring loop.

\subsection{Atom-to-token pooling and the determinism boundary}
\label{app:2d-pooling}

The 1D scheme requires a rank-$0$ \textsc{Broadcast} of $\mathbf{A}_I$ immediately after \texttt{process\_a} (Appendix~\ref{app:nondet}), because \texttt{torch.Tensor.index\_reduce("mean")} is non-deterministic across devices. The 2D scheme handles the same boundary in a structurally different way: the atom$\to$token pooling is implemented as a \texttt{DistributedScatterReduce} collective along $\text{cp}_0$, with the per-element scatter indices coming from the (already replicated) global \texttt{tok\_idx}. Because the reduction is performed by a single deterministic collective rather than by per-device atomic accumulators, the resulting $\mathbf{A}_I$ is bit-identical across ranks by construction and no separate broadcast is needed. The same primitive also implements the irregular C$\alpha$/motif-token pooling consumed by the distogram path of the diffusion token encoder.

\subsection{Sampling-loop integration}
\label{app:2d-loop}

The ring primitive is invoked inside every recycling iteration of every denoising step, identically to how the 1D scheme invokes its per-block cross-attention. At diffusion-step boundaries, rank~$0$ broadcasts the current noised coordinates, the freshly sampled Gaussian noise, and the denoised prediction; symmetrisation, when active, is then applied independently on every rank using the broadcast inputs, so that the trajectory stays deterministic without any extra mesh-aware bookkeeping. Because the 1D and 2D schemes share the same step-boundary broadcast pattern, they see the same stochastic trajectory for a given seed, which simplifies head-to-head comparisons of correctness and timing.

\subsection{2D-specific data-pipeline transforms}
\label{app:2d-transforms}

Two pre-pipeline transforms are introduced for 2D CP to avoid materialising dense $[I, I]$ matrices in the data loader. Both store a small 1D representation in the feature dictionary and defer the reconstruction of the local $[I_r, I_c]$ quadrant to the model's input layer, where the quadrant is built directly as a DTensor on the resident grid position. \texttt{CPAwareUnindexFlaggedTokens} replaces the standard $[I, I]$ unindexing pair mask by a pair of $[I]$-shaped tensors -- a per-token \texttt{is\_unindexed} boolean and a \texttt{group\_ids} integer assignment -- from which the quadrant of the mask is recomputed on-rank. \texttt{AddAF3TokenBondFeatures} replaces the dense $[I, I]$ token-bond matrix by a sparse COO representation when the structure exceeds a configurable atom-count threshold (default $5\times10^{4}$), again reconstructing the quadrant on-rank inside \texttt{CPTokenInitializer}. Together, these transforms keep the data loader's memory footprint bounded by $\mathcal{O}(I)$ even for the largest assemblies we target, where the dense matrices would already exceed several hundred megabytes per sample.

\subsection{Per-component memory optimisations carried over to 2D}
\label{app:2d-memory}

Several of the optimisations of Appendix~\ref{app:memory} carry over to the 2D scheme; others are subsumed by the inherently quadrant-based layout. Specifically:
\begin{itemize}
\item $\mathbf{Z}$ striping and $\mathbf{D}_\text{self}$ chunking are not separate optimisations on 2D: every shape that the 1D scheme writes as $[I/P, I, \cdot]$ exists on 2D as a $[I_r, I_c, \cdot]$ DTensor by construction.
\item $\mathbf{P}$ sparsification reuses the same chunked pairwise embedder as 1D, with the static MLP projections cached once at tokenisation; the only difference is that on 2D, the per-rank slice of $\mathbf{P}$ is a quadrant rather than a row stripe.
\item $\mathbf{Z}_\text{aug}$ pre-allocation in place of \texttt{cat} is again applied to the diffusion-token-encoder concatenation, since the channel-dimension cat is the same regardless of whether the leading two dimensions are sharded as $[I/P, I, \cdot]$ or $[I_r, I_c, \cdot]$.
\item Manual SwiGLU decomposition and the explicit $4{\times}$ RPE sub-chunking from the 1D scheme are not currently applied on 2D, because the local per-rank $\mathbf{Z}$ quadrant is small enough that the unmodified Fold-CP transition and RPE primitives stay below the per-GPU memory budget at the assembly sizes we target. They could be ported across schemes if a future 2D configuration moved the bottleneck back to the transition or RPE blocks.
\end{itemize}

\subsection{Class structure and environment variables for the 2D scheme}
\label{app:2d-classes}

The 2D entry point is \texttt{create\_rfd3\_distributed(rfd3, manager)}, which is invoked by the engine after model construction and after the \texttt{DistributedManager} has produced its mesh and subgroup layouts. The helper instantiates the four CP wrapper modules listed in Appendix~\ref{app:2d-mesh}, attaches their \texttt{forward} methods to the corresponding serial submodules, redistributes the parameters via \texttt{distribute\_params}, and runs the post-construction DTensor validation check.

The 2D scheme reuses \texttt{RFD3\_ATTENTION\_PARALLEL} as its top-level on/off switch and inherits \texttt{RFD3\_LOW\_MEMORY\_MODE} (which controls the chunked pairwise embedder shared with 1D) and \texttt{RFD3\_NCCL\_TIMEOUT\_SEC} (the elevated NCCL watchdog timeout). The atom-level inter-chain attention budget is configured through three additional optional variables -- \texttt{RFD3\_N\_ATTN\_KEYS}, \texttt{RFD3\_INTER\_CHAIN\_WEIGHT}, and \texttt{RFD3\_ATOM\_USE\_ICA} (with companion \texttt{RFD3\_ATOM\_INTER\_CHAIN\_WEIGHT}) -- and two debug-oriented variables -- \texttt{RFD3\_DEBUG\_STATS} and \texttt{RFD3\_MEM\_PROFILE} -- toggle a checkpoint-statistics logger and a per-step memory profile, respectively. These last five variables are not strictly part of the CP machinery but are exposed by the same code path because they affect the per-block cost of the ring attention and are therefore relevant for reproducing the timings reported in \S\ref{sec:results-scaling}.


\section{Designability metrics}
\label{app:designability}

This appendix provides the formal definitions of the \emph{in silico} metrics used in \S\ref{sec:results-quality} and \S\ref{sec:results-democratization}. All quantities are computed directly from the generated all-atom coordinates of the symmetrised assembly; no external structure-prediction oracle is required. We split the metrics into two families: backbone sanity (per-design checks on the diffused chain itself) and symmetric-interface sanity (checks that probe the inter-subunit geometry of the assembly).

\paragraph{Backbone sanity.} These metrics flag designs whose monomeric chain is geometrically broken, irrespective of any symmetry consideration.
\begin{itemize}
  \item \textbf{Chain breaks (\texttt{n\_chainbreaks}).} For every consecutive pair of C$\alpha$ atoms along the chain, we compute the bond length and its deviation from the canonical $3.8$\,\AA{}; pairs with deviation above $\tau_\text{cb}=0.75$\,\AA{} are flagged as chain breaks. Pairs that span an intentional chain transition (different \texttt{chain\_iid}) are masked out so they do not contribute. We additionally report \texttt{max\_ca\_deviation}, the worst per-design deviation, as a continuous summary. A geometrically clean monomer should have \texttt{n\_chainbreaks}~=~0.
  \item \textbf{Inter-residue clashes (\texttt{n\_clashing.interresidue\_clashes\_w\_backbone} and \texttt{...\_w\_sidechain}).} We pairwise compare all heavy atoms of the protein and count pairs that (i) belong to residues at least two apart along the sequence and (ii) are closer than $\tau_\text{clash}=1.5$\,\AA{}. The backbone-only variant restricts the second factor to atoms in $\{\text{N},\text{C}\alpha,\text{C}\}$ and is the more conservative indicator of physical implausibility because the backbone has no rotameric flexibility to relieve the clash. The sidechain variant is much noisier and is reported only as a lower bound on clash density.
  \item \textbf{Non-loopy fraction (\texttt{non\_loop\_fraction}).} We run the P-SEA secondary-structure annotator~\cite{labesse_p-sea_1997} (as implemented in Biotite's \texttt{annotate\_sse}) on the diffused chain and report the fraction of residues assigned to helix~or~strand, i.e.\ the complement of the coil fraction.
\end{itemize}

\paragraph{Symmetric-interface sanity.} These metrics use the full symmetrised complex and exclude any atoms with \texttt{sym\_transform\_id}~$<0$ (fixed/unsymmetrised motifs). All distances are C$\alpha$--C$\alpha$. We use three thresholds that come from the implementation in \texttt{rfd3.inference.symmetry.metrics}: a hard inter-subunit clash distance $\tau_\text{clash}^\text{sym}=3.5$\,\AA{}, an interface-contact band $[\tau_\text{contact}^\text{lo},\tau_\text{contact}^\text{hi}]=[4.0,10.0]$\,\AA{}, and a proximity cutoff $\tau_\text{prox}=15.0$\,\AA{} above which two subunits are considered non-interacting.
\begin{itemize}
  \item \textbf{Minimum inter-chain distance (\texttt{complex.min\_inter\_chain\_distance}).} The smallest C$\alpha$--C$\alpha$ distance between atoms belonging to two distinct subunits anywhere in the complex. Values below $\sim 3.5$\,\AA{} indicate steric overlap; values much above $\sim 6$\,\AA{} indicate that the asymmetric units never actually meet, which for a closed nanoparticle is a failure mode.
  \item \textbf{ASU clashes (\texttt{asu.n\_clashes}).} The number of inter-subunit C$\alpha$--C$\alpha$ pairs closer than $\tau_\text{clash}^\text{sym}$, normalised by the number of subunits. The normalisation makes the metric directly comparable across symmetries with different oligomeric states.
  \item \textbf{Mean contacts per interface (\texttt{complex.mean\_contacts\_per\_interface}).} For each pair of subunits whose C$\alpha$--C$\alpha$ atoms come within $\tau_\text{prox}$ of each other we count the number of C$\alpha$--C$\alpha$ pairs falling inside the contact band $[\tau_\text{contact}^\text{lo},\tau_\text{contact}^\text{hi}]$. We then average this count over all proximal interfaces. Higher values indicate richer, better-formed interfaces.
  \item \textbf{Total interface contacts (\texttt{complex.n\_interface\_contacts}).} The unnormalised sum of the above over all interfaces.
  \item \textbf{Number of interfaces with contacts (\texttt{complex.n\_interfaces\_with\_contacts})} and \textbf{minimum contacts per interface (\texttt{complex.min\_contacts\_per\_interface}).} These flag assemblies in which one of the interfaces is essentially absent (low minimum) even when the mean is healthy.
\end{itemize}

\paragraph{Additional secondary-filter metrics.} We additionally compute the radius of gyration of the diffused chain (Biotite's \texttt{gyration\_radius}); the helix, sheet, and loop fractions and the number of secondary-structure elements from the same P-SEA annotation; per-residue amino-acid composition (in particular alanine and glycine content, where biased composition often signals a pathological design); and the smallest C$\alpha$--C$\alpha$ distance \emph{within} the ASU (\texttt{asu.min\_intra\_distance}), which catches intra-subunit overlap that is invisible to the inter-subunit clash count.

\paragraph{Protocol.} Unless otherwise stated, the figures in \S\ref{sec:results-quality} aggregate over $n=40$ independent designs targeting an icosahedral assembly with $210$ residues per chain, generated with the 2D sharding scheme on a $2\times 2$ device grid of HG200 GPUs ($95$\,GB each) and using a batch size of one.

\section{Per-chain comparison: Design-CP vs vanilla RFD3 (icosahedral)}
\label{app:vanilla-rfd3-vs-designcp}

\begin{figure*}[!t]
  \centering
  \includegraphics[width=\textwidth]{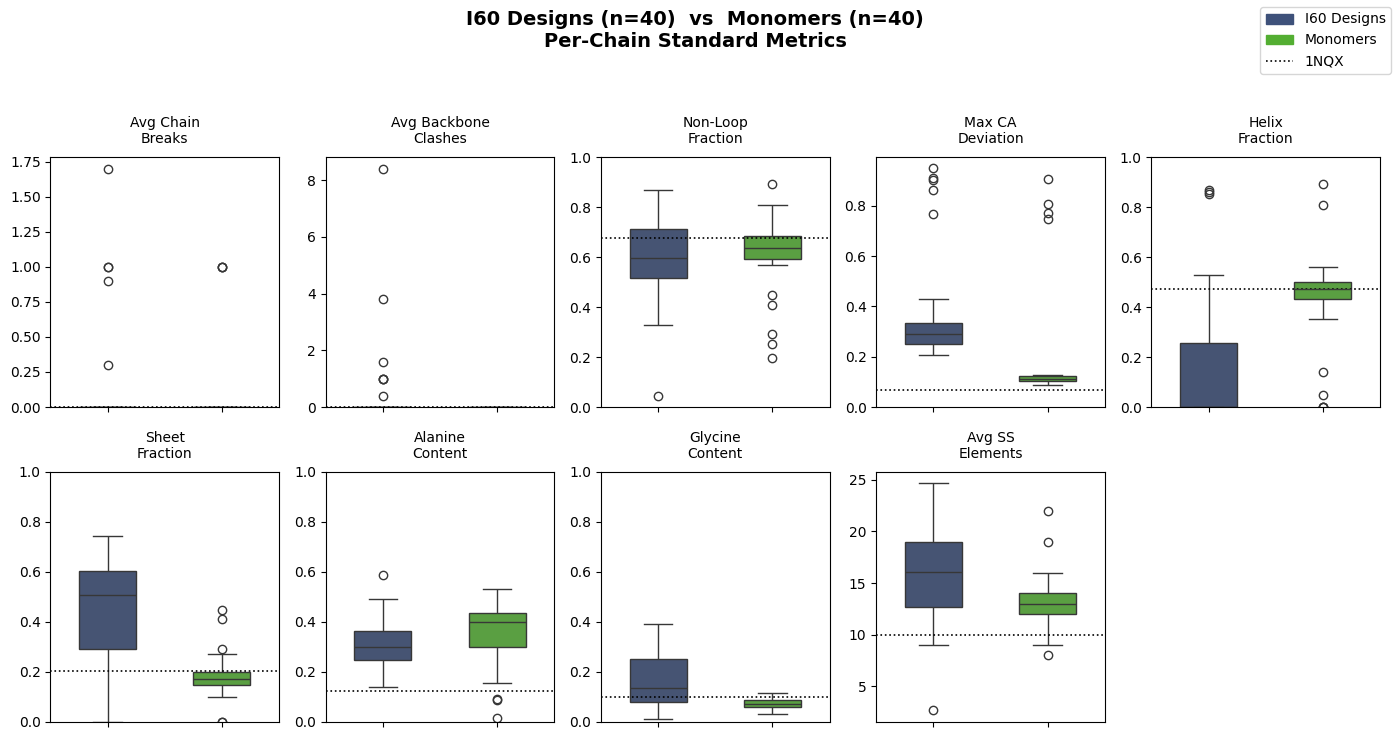}
  \caption{\textbf{Per-chain comparison of Design-CP icosahedral designs against vanilla RFD3 monomers (full eight panels).} Distributions of standard backbone-sanity and composition metrics, computed chain-by-chain on $n=40$ Design-CP icosahedral assemblies (blue, $60$ chains $\times$ $210$ residues per chain) and $n=40$ vanilla single-GPU RFD3 monomers of length $210$ (green). The dotted line in each panel reports the corresponding value for Lumazine Synthase (PDB:~1NQX). The first four panels reproduce \cref{fig:sym-neighbourhood-quality}b--e and are commented on in the main text; the additional four panels (helix fraction, sheet fraction, glycine content, average number of secondary-structure elements) reveal the secondary-structure shift discussed in this appendix.}
  \label{fig:i60-vs-monomers}
\end{figure*}

The main-text comparison in \cref{fig:sym-neighbourhood-quality}b--e covers four panels (chain breaks, backbone clashes, non-loop fraction, max CA deviation). \Cref{fig:i60-vs-monomers} shows the full eight-panel version, adding helix fraction, sheet fraction, glycine content, and average number of secondary-structure elements per chain.

\paragraph{Caveat on problem difficulty.} The two design problems are not of equal difficulty and the comparison should be read with this asymmetry in mind. Each icosahedral chain is denoised inside a $12{,}600$-residue joint context where its trajectory must remain self-consistent with the simultaneously denoised trajectories of $59$ symmetry mates and with all the inter-chain interfaces they form, while the monomer problem is a single $210$-residue chain comfortably inside RFD3's native $384$-token training crop. The icosahedral problem is therefore strictly harder per chain.

\paragraph{Helix and sheet fraction.} One of the most prominent qualitative gap in \cref{fig:i60-vs-monomers} appears in secondary-structure composition. Vanilla RFD3 monomers reproduce the helix/sheet balance similar to the one of our reference structure 1NQX: the helix fraction is concentrated around a median of $\approx 0.47$ with a tight bulk against the 1NQX reference of $0.47$, and the sheet fraction sits at a median of $\approx 0.18$ with a comparable spread, against $0.20$ for 1NQX. Design-CP icosahedral designs, in contrast, are strongly $\beta$-biased: the helix fraction collapses near zero on the bulk of chains with a few upper outliers and a handful of smaller ones, and the sheet fraction shifts upwards to a median of $\approx 0.5$ with a noticeably broader distribution.

\paragraph{Number of secondary-structure elements.} Consistent with the helix/sheet shift, the average number of secondary-structure elements per chain rises from $\approx 13$ in the monomers to $\approx 16$ in the icosahedral designs, with the icosahedral distribution carrying a heavier upper tail. Both populations include outliers, but only the icosahedral set reaches the upper-twenties range. At fixed chain length, a higher element count mechanically implies shorter average element length, which is again consistent with the icosahedral chains preferring multiple short $\beta$-strands over a small number of long helices. We treat this as supportive evidence for the helix/sheet shift rather than an independent observation, and note that the metric is sensitive to the P-SEA assignment thresholds~\cite{labesse_p-sea_1997}.

\paragraph{Amino-acid composition.} On amino-acid composition the two populations are closer to each other than on secondary structure. Alanine content is similar across both sets, with broadly overlapping distributions and medians on the same order, and is, as is typical for RFD3~\cite{butcher_novo_2025}, slightly inflated relative to 1NQX in both cases; we do not read a meaningful difference between Design-CP ASUs and vanilla monomers on this axis. Glycine content, in contrast, is appreciably broader in the icosahedral designs ($\approx 0.10$--$0.25$, with sporadic high outliers) than in the monomers ($\approx 0.05$--$0.10$, very tightly concentrated), although the medians remain on the same order and within the typical range for natural proteins. The widened glycine spread is qualitatively consistent at the population level with the broader max-C$\alpha$--C$\alpha$ deviation distribution of \cref{fig:sym-neighbourhood-quality}e -- chains under more inter-subunit geometric strain might rely more on backbone-flexible residues -- but we flag this as a tentative association rather than a causal claim, since we have not tested whether the same chains drive both effects.

\section{Octahedral design metrics on commodity GPUs}
\label{app:O24_scaling_and_metrics}

\begin{figure}[!t]
  \centering
  \begin{subfigure}[b]{\linewidth}
    \centering
    \includegraphics[width=0.87\linewidth]{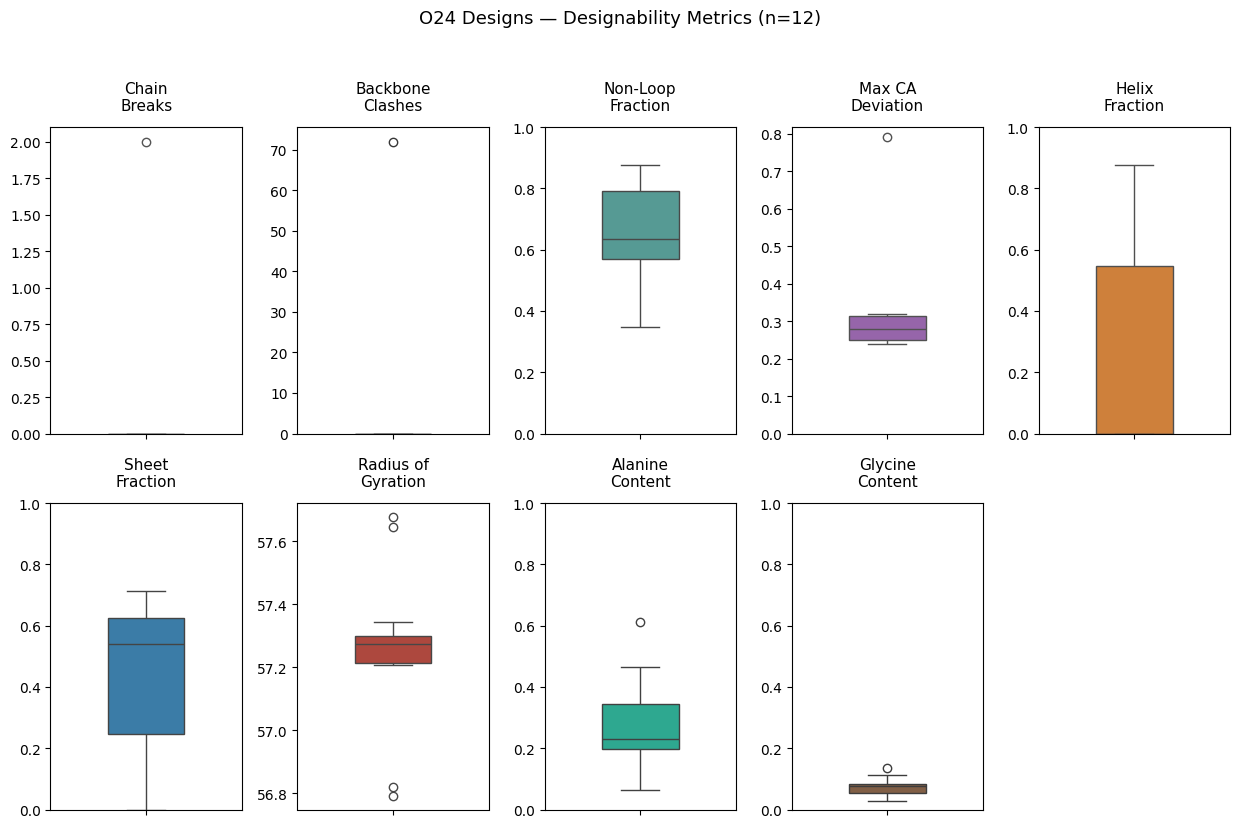}
    \label{fig:o24-designability}
  \end{subfigure}
  \\[0.2em]
  \begin{subfigure}[b]{\linewidth}
    \centering
    \includegraphics[width=0.73\linewidth]{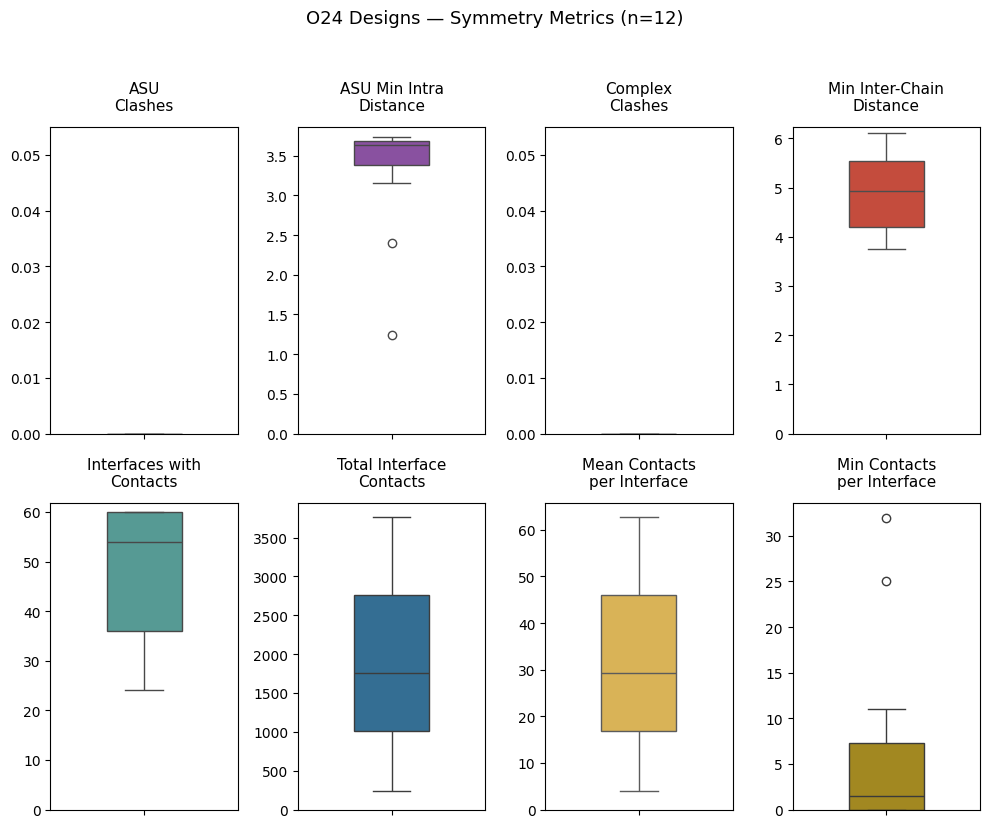}
    \label{fig:o24-symmetry}
  \end{subfigure}
  \caption{\textbf{Full distributions of \emph{in silico} metrics for octahedral designs ($n=12$).} Headline metrics from \cref{fig:octahedral-mainfig}b--g are reproduced here together with the additional panels discussed in this appendix. \textbf{(a)}~Backbone-sanity metrics: chain breaks, backbone clashes, non-loop fraction, max CA deviation, helix fraction, sheet fraction, radius of gyration, alanine content, glycine content. \textbf{(b)}~Symmetry-interface metrics: ASU clashes, ASU minimum intra-distance, complex clashes, minimum inter-chain distance, number of interfaces with contacts, total interface contacts, mean contacts per interface, minimum contacts per interface.}
  \label{fig:o24-all}
\end{figure}

This appendix reports the full distributions of \emph{in silico} metrics for the $n=12$ octahedral assemblies generated on $16$ NVIDIA RTX~A4000 GPUs ($16$\,GB each) with a batch size of one. The headline metrics (chain breaks, backbone clashes, non-loop fraction, max CA deviation, minimum inter-chain distance, mean contacts per interface) are already shown in \cref{fig:octahedral-mainfig}b--g and discussed in \S\ref{sec:results-democratization}. \Cref{fig:o24-all} shows the full set of backbone-sanity and symmetry-interface metrics; we comment below only on the additional panels that are not in the main text.

\paragraph{Additional backbone-sanity panels (\cref{fig:o24-all}a).} The radius of gyration is tightly concentrated in the $\approx 56.8$--$57.4$\,\AA{} range, indicating a consistent per-chain envelope across the population. Helix and sheet fractions are both broad, with no obvious dominant secondary-structure type within the population: helix fraction spans $\approx 0$--$0.9$ with a median around $\approx 0.55$, and sheet fraction spans $\approx 0$--$0.85$ with a median around $\approx 0.45$. This is in qualitative contrast with the icosahedral designs of Appendix~\ref{app:vanilla-rfd3-vs-designcp}, where helix fraction collapses near zero. We interpret this as plausibly reflecting the lower oligomeric state ($24$ vs.\ $60$ subunits), which gives the symmetrisation step less reason to favour extended $\beta$-sheets over helical packing, but caution that the sample size ($n=12$) is small. Alanine content has a median of $\approx 0.25$ and is comparatively broad ($\approx 0.05$--$0.6$), while glycine content is tightly clustered around $\approx 0.05$--$0.10$. These figures are in line with what observed in \cref{app:vanilla-rfd3-vs-designcp}.

\paragraph{Additional symmetry-interface panels (\cref{fig:o24-all}b).} The hard inter-subunit clash counts at both the ASU and complex levels (\texttt{asu.n\_clashes} and \texttt{complex} clashes) are at zero across the entire population, indicating that no design realises sterically forbidden inter-subunit contacts. The smallest intra-ASU C$\alpha$--C$\alpha$ distance is centred at $\approx 3.6$\,\AA{} (matching the canonical C$\alpha$--C$\alpha$ spacing) with two low outliers at $\approx 1.2$ and $\approx 2.4$\,\AA{} that flag designs with intra-ASU geometric strain. The number of interfaces showing contacts has a median of $\approx 55$ out of the $24$ chains' interface budget, and the total number of interface contacts spans $\approx 200$--$3{,}700$ with a median of $\approx 1{,}700$. The minimum number of contacts per interface has a median near $1$, with two outliers at $\approx 25$ and $\approx 30$; this reflects that some designs do contain weak interfaces whose contact count drags the per-design minimum near zero, a useful filter signal for downstream design campaigns.

Across both metric families, the additional panels are consistent with the headline result of \S\ref{sec:results-democratization}: octahedral designs sampled on commodity GPUs satisfy the same hard-failure criteria as the icosahedral baseline, with the main qualitative difference being a more permissive secondary-structure distribution.

\section{Effect of removing the symmetry constraint}
\label{app:asymmetric_multi_chains}

\begin{figure*}[!t]
  \centering
  \includegraphics[width=\textwidth]{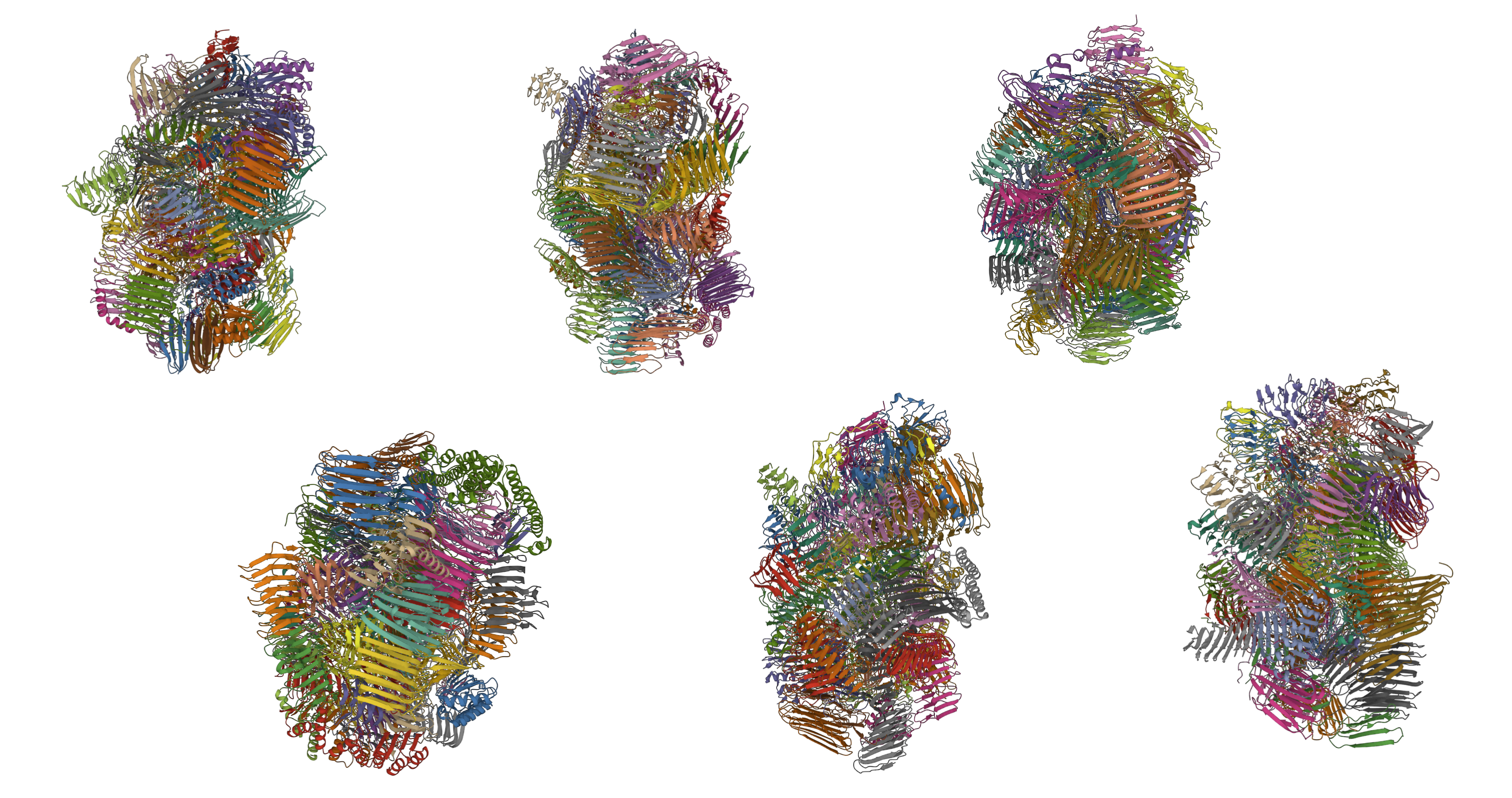}
  \caption{\textbf{Effect of removing the symmetry constraint at the same system size.} Six Design-CP samples generated with $60$ chains of $210$ residues each ($12{,}600$ residues total) under \emph{no} point-group symmetry constraint, with each colour denoting a distinct chain. Compare with the well-formed icosahedral nanoparticles of \cref{fig:sym-neighbourhood-quality}f.}
  \label{fig:monstromers}
\end{figure*}

The main text argues (\S\ref{sec:results-symmetry}) that strong point-group symmetry constraints are what make Design-CP usable on system sizes well beyond RFD3's native $384$-token training crop. \Cref{fig:large-monomer-quality}b already illustrated this for the extreme case of a single $10{,}800$-residue monomer. \Cref{fig:monstromers} shows the complementary illustration for the multi-chain regime relevant to nanoparticle design: six Design-CP samples generated with exactly the same total system size as the icosahedral assemblies of \S\ref{sec:results-quality} ($60$ chains of $210$ residues each, $12{,}600$ residues in total) but with \emph{no} point-group symmetry constraint imposed at sampling time. Token and atom counts, the architecture, the weights, and the parallelisation strategy are kept identical to the symmetric runs; only the ASU restriction described in Appendix~\ref{app:symmetry} is removed.

The resulting structures are visibly degenerate: large slabs of $\beta$-strands packed in irregular orientations, no recognisable globular folds, and no consistent inter-chain organisation. None of these samples resemble naturally occurring multimeric proteins, and they bear no resemblance to the well-formed icosahedral assemblies in \cref{fig:sym-neighbourhood-quality}f despite using exactly the same number of tokens, atoms, and chains. We take this as direct qualitative evidence that the dominant factor behind the sample-quality results in \S\ref{sec:results-quality} is the symmetry prior rather than the lenght of the modelled chains: at this scale, RFD3 + Design-CP cannot recover plausible protein-like geometry from joint denoising alone.

\end{document}